%% file: main.tex
\pdfoutput=1
\documentclass[conference]{IEEEtran}
\IEEEoverridecommandlockouts
\usepackage{cite}
\usepackage{amsmath,amssymb,amsfonts}
\usepackage{graphicx}
\usepackage{textcomp}
\usepackage{xcolor}
\usepackage{booktabs}
\usepackage[hyphens]{url}

\usepackage[bookmarks=true, pdfstartview=FitH]{hyperref}

\newcommand{\sys}{LoomKV}
\input{sections/pathstep}

\begin{document}
\bstctlcite{IEEEexample:BSTcontrol}

\title{LoomKV: Lossless KV Cache Compression from GPU Memory to SSD for LLM Serving}

\author{
\IEEEauthorblockN{
Wentao Zhu$^{\dagger}$,
Ziao Sheng$^{\dagger}$,
Peng Deng$^{\dagger}$,
Huiqi Hu$^{\dagger\ast}$\thanks{$^\ast$Corresponding author.},
Xuan Zhou$^{\dagger}$,
Xing Wei$^{\ddagger}$,
Zhuang Wang$^{\ddagger}$}
\IEEEauthorblockA{$^{\dagger}$\textit{East China Normal University}, $^{\ddagger}$\textit{ZTE Corporation}\\
$^{\dagger}$\{wtzhu, 52315903002, 52295903003\}@stu.ecnu.edu.cn, $^{\dagger}$\{hqhu, xzhou\}@dase.ecnu.edu.cn\\
$^{\ddagger}$\{wei.xing6, wang.zhuang\}@zte.com.cn}
}

\maketitle

\input{sections/00-abstract}

\input{sections/01-introduction}

\input{sections/02-background}

\input{sections/03-overview}
\input{sections/04-codec}
\input{sections/05-lmcache}
\input{sections/06-gpu-pool}
\input{sections/06s-sync}
\input{sections/08-evaluation}
\input{sections/09-mode-choice}
\input{sections/11-related-work}
\input{sections/12-conclusion}

\bibliographystyle{IEEEtran}
\IEEEtriggeratref{33}
\bibliography{references}

\end{document}

%% file: sections/pathstep.tex
\newcommand{\pathstep}[1]{%
  \begingroup
  \setlength{\unitlength}{1pt}%
  \begin{picture}(11,8.5)%
    \put(5.5,3.4){\circle{10}}%
    \put(5.5,3.4){\makebox(0,0){\normalfont\sffamily\bfseries\scriptsize #1}}%
  \end{picture}%
  \endgroup}

%% file: sections/00-abstract.tex
\begin{abstract}
Large language model (LLM) serving engines offload the key/value (KV) cache
from GPU memory to host DRAM and SSD so multi-turn and agent requests can reuse
prompt prefixes. Capacity then limits service at two levels—external tiers
govern prefix reuse before eviction, while the GPU KV pool bounds concurrent
long requests—and lossy compression alters KV values and outputs. \sys{} brings
lossless KV compression to both external tiers (\emph{tier mode}) and the GPU
pool (\emph{pool mode}) on an unmodified engine. Using a calibrated per-head
two-level 3-bit code over BF16 high bytes (sign and upper seven exponent bits),
\sys{} stores BF16 KV at 0.71 of its raw size across nine models and FP8 KV at
0.85--0.87. In tier mode, \emph{TileStream} writes exact-length records with
length readback off the load path—leaving tier storage, eviction, and prefetch
unchanged—and decompresses fetched records straight into scattered GPU pages.
In pool mode, \emph{FragPack} packs variable-length records into smaller
shared-budget GPU pages read in place by a fused decode-attention kernel faster
than FlashAttention-3 on BF16. Built with LMCache and vLLM, \sys{} makes
compressed tiers act as 1.4$\times$ larger raw tiers on AgentX: across
LMCache's LRU cliff, tier mode delivers 2.7--3.6$\times$ raw's goodput and
1.9--3.1$\times$ lower mean time-to-first-token (TTFT), shifts the cliff
identically under other simulated replacement policies, and cuts
repeated-prompt TTFT up to 1.7$\times$ via SSD-to-DRAM promotion. When tiers
hold the working set, pool mode fits 1.31--1.34$\times$ as many GPU tokens,
admitting 12 long requests with roughly raw's queue and goodput at 8. On a production-derived AI IDE trace where both tiers and the GPU pool are
constrained, tier and pool modes raise goodput by 1.24$\times$ and
1.59$\times$. Where raw capacity suffices, compression costs
at most 2.2\% throughput; tier mode is bit-exact with raw tiers, and pool mode
matches FlashAttention-3's numerical accuracy.
\end{abstract}


%% file: sections/01-introduction.tex
\section{Introduction}
\label{sec:intro}
Large language model (LLM) requests often share long prompt prefixes. An agent re-sends its growing
transcript on every turn, while chat and retrieval-augmented (RAG) requests
repeat history and documents~\cite{kvwild,copilot}. Processing these prefixes from scratch would repeat prefill work.
Serving engines therefore retain their key/value (KV) cache so that returning requests can skip this computation. That cache
is large. For instance, Qwen3-30B-A3B stores 96\,KB of KV per token, and a GPU that
reserves 32\,GB for KV holds only five 64K-token agent sessions. Engines
such as vLLM~\cite{vllm} and SGLang~\cite{sglang} therefore spill KV to host
DRAM and SSD through external cache tiers (LMCache~\cite{lmcache},
Mooncake~\cite{mooncake}, SGLang HiCache~\cite{hicache},
CachedAttention~\cite{cachedattention}, Strata~\cite{strata}) and load it
back when a prefix returns.

Cache capacity affects serving performance at two levels: the external
DRAM/SSD cache tiers and the GPU KV pool (\S\ref{sec:bg-cap}).
First, external cache capacity determines how much previously computed KV
can be retained for reuse. When needed KV has been evicted, the engine must
recompute it through prefill instead of loading it from cache. This increases
computation and request latency.
Second, GPU KV capacity limits how many long requests can run concurrently.
Each active request needs space for its KV in the GPU pool. Even if its
prefix is cached in DRAM or SSD, a request must wait when the GPU pool is
full. Increasing external cache capacity does not provide this GPU space.

Thus, cache capacity matters at every level of the hierarchy. Storing KV in
fewer bytes would add capacity at both places at
once, without adding memory or drives. Most KV compression does so lossily: it quantizes values
(KIVI~\cite{kivi}, FP8 KV, CacheGen~\cite{cachegen}, KVTC~\cite{kvtc}) or evicts tokens predicted to matter little (H$_2$O~\cite{h2o}, SnapKV~\cite{snapkv}). Both
change outputs, and accuracy must then be requalified per model and task.

Lossless compression can increase effective cache capacity without changing
any KV value. In Qwen3-30B-A3B, the seven most frequent BF16 high-byte
values per K or V tensor account for about 94\% of high bytes on average
(Fig.~\ref{fig:motiv}c). Because lossless compression preserves the
stored representation exactly, it can also be applied to quantized formats
such as FP8 KV.

We present \sys{}, which applies lossless compression to
both the external cache tiers and the GPU KV pool.
\sys{} supports two modes: \emph{tier mode} and \emph{pool mode}.
In \emph{tier mode}, DRAM and SSD store compressed KV, which is restored to
its original form in the GPU pool before vLLM's existing attention backend
uses it. In \emph{pool mode}, the GPU pool also stores compressed KV, and a
custom decode-attention kernel reads and decompresses its compressed records as it
computes attention.

In both modes, the question is whether the saved bytes become capacity. A
compressed tier or pool holding $B$ bytes at stored ratio $r$ has $B/r$
\emph{raw-equivalent} bytes. We call it \emph{capacity-equivalent} if it
performs like a raw one of that size.
Achieving this equivalence requires preserving the space savings while
keeping the overhead of compression and decompression low. This leads to
five design requirements:

(i) \emph{Compression ratio:} BF16 KV has limited redundancy, so the potential
space savings from lossless compression are modest. Headers, indexes, and
alignment padding must therefore be kept small to preserve those savings.

(ii) \emph{Reducing allocated space:} in both modes, compression must
reduce the memory or storage space allocated to the same amount of KV,
so that the saved space can hold additional KV. Reserving the original
uncompressed space for that KV would leave cache capacity unchanged.

(iii) \emph{Fast parallel decompression in tier mode:} inference frameworks
such as vLLM manage GPU KV in fixed-size pages, and a request's pages need
not be contiguous in GPU memory. When loading compressed KV from external
cache tiers, decompression must exploit GPU parallelism to restore the
original values directly into these pages at high throughput, avoiding an
extra copy through a temporary buffer.

(iv) \emph{Fast attention in pool mode:} each decode step must read KV
directly from compressed GPU pages and recover the values needed for attention.
The combined decompression and attention computation must match the speed
of the engine's existing kernels, FlashAttention-3 (FA3)~\cite{fa3} and
FlashInfer~\cite{flashinfer}.

(v) \emph{Avoiding synchronization stalls:} when storing compressed KV in
external cache tiers, the CPU must read the compressed KV sizes back
from the GPU to allocate space and determine transfer sizes. This
readback requires synchronization, but the wait must not block other KV
cache loads.

We design \sys{} to meet these five requirements. Our main contributions are:
\begin{itemize}
\item \textbf{A lossless KV compression code}
(\S\ref{sec:codec}; requirements~i--ii): we design a lossless KV code in the same encoding
family as ZipServ~\cite{zipserv}.
With calibrated codebooks, its tier-mode records use about 71\% of the original
BF16 KV storage across nine evaluated models, while preserving every KV
value. To turn these savings into usable capacity, external tiers give each tier-mode record an object of its exact length, while the GPU pool packs variable-length pool-mode records into smaller fixed-size physical pages.
\item \textbf{Parallel GPU decompression in tier mode}
(\S\ref{sec:lmc}; requirement~iii): building on this
encoding scheme, we design TileStream, a compressed KV layout that
organizes bytes by type into separate data streams. Per-tile indexes locate
the variable-length portions of these
streams. Our GPU decompressor uses these indexes to decompress tiles in
parallel, one per warp, and writes restored KV directly into the engine's scattered
cache pages, avoiding an intermediate copy.
\item \textbf{Fused decompression and decode attention}
(\S\ref{sec:hbm}, requirement~iv): in pool mode, we design the FragPack layout
and a kernel that fuses decompression with decode attention. FragPack packs
compressed KV values in the order GPU threads need to form their operand
fragments for matrix multiplication. Using this layout, the fused kernel
decompresses each thread's KV values directly into registers and
immediately uses them in attention's matrix multiplications, avoiding a
separate pass that writes uncompressed KV back to GPU memory.
\item \textbf{Avoiding synchronization stalls in KV transfers}
(\S\ref{sec:sync}; requirement~v): we design a method
that moves the CPU's readback of GPU-computed lengths off the critical
path when writing variable-length compressed KV to external cache tiers.
This prevents the synchronization required for length readback from
blocking other KV loads.
\item \textbf{Guidance for choosing a compression mode}
(\S\ref{sec:eval}, \S\ref{sec:disc}): based on our evaluation of both
modes, we give recommendations for selecting tier or pool mode using
observations of system state. These
recommendations help target the capacity bottleneck and avoid adding
capacity where it does not improve serving performance.
\end{itemize}

We implement \sys{} on unmodified vLLM, with tier mode inside the
cache server of LMCache~\cite{lmcache} and pool mode as an out-of-tree
attention backend. LMCache continues to manage allocation, eviction,
write-through, and prefetch.
We evaluate the two places where capacity limits service in turn, on the agent traces of AgentX~\cite{agentx}. First, tier mode achieves a compression ratio of about 71\% for BF16 KV (compressed size divided by raw size).
With 880\,GB of DRAM and SSD combined, it matches the goodput of a raw deployment with 1231\,GB.
At the cliff of LMCache's write-through LRU tiers, where raw tiers evict prefixes just before reuse, tier mode serves 2.7--3.6$\times$ raw's goodput. Other policies move this cliff (\S\ref{sec:disc}). Second, once the tiers hold the
working set, a compressed GPU pool is capacity-equivalent at its token count:
admitting 12 long requests at once, it has about the queue and goodput that a raw pool
has with 8. Capacity also
helps time-to-first-token (TTFT): hits move to faster tiers, cutting the TTFT of repeated prompts
up to 1.7$\times$.
The cost is up to 2.2\% of throughput where raw tiers suffice and about 2.5\,GB
of GPU memory per GPU in tier mode, and tier mode's outputs are those of raw
tiers, bit for bit.

%% file: sections/02-background.tex
\section{Background and Motivation}
\label{sec:bg}
\subsection{KV cache management}
The KV cache hierarchy spans GPU memory and external cache tiers.
Inference frameworks commonly manage GPU KV memory as a pool of fixed-size
pages. Pages are allocated as requests' KV grows and reclaimed when no
longer needed. A page table maps each request's logical token blocks to
physical pages, allowing attention to access KV stored at noncontiguous
GPU addresses.
External cache tiers retain reusable KV in DRAM ($L_1$) and SSD ($L_2$).
When a prompt prefix is found in an external tier, its KV is loaded into
GPU pages to avoid recomputing that prefix. These tiers manage allocation,
eviction, and data transfers to preserve KV for reuse across requests.

\subsection{Where capacity limits service}
\label{sec:bg-cap}
\begin{figure*}[t]
\centering
\includegraphics{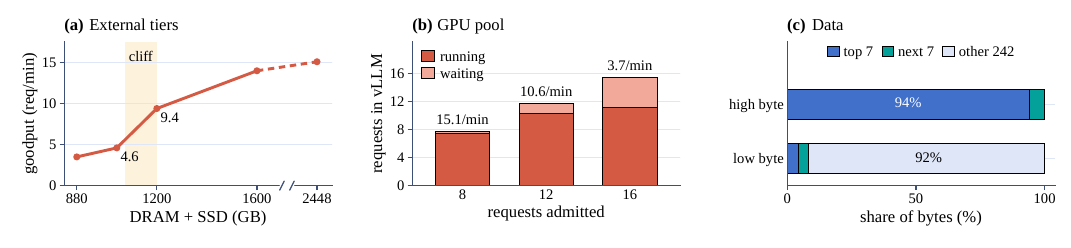}
\caption{Three observations on raw KV. (a) Goodput by tier size.
(b) Requests running and waiting in vLLM. (c) Mean shares of BF16 high and
low bytes covered by the top 7, next 7, and remaining 242 values. We rank
high- and low-byte values separately within each tensor and average the
shares over all 96 tensors (one K and one V per layer across 48 layers).}
\label{fig:motiv}
\end{figure*}

For the measurements in Fig.~\ref{fig:motiv}, we serve Qwen3-30B-A3B on
two H800 GPUs using vLLM and LMCache, with KV stored in uncompressed BF16.
The AgentX workload consists of 64 live sessions
sharing a 400\,GB DRAM cache.

Panel (a) varies SSD capacity. Goodput counts requests per minute with engine time to first
token $\le$ 2\,s and inter-token latency $\le$ 40\,ms. Reducing total
DRAM and SSD capacity from 1200\,GB to 1040\,GB nearly halves goodput,
from 9.4 to 4.6 requests per minute, as cached KV is evicted before reuse.
This capacity cliff contrasts with the smaller gains at larger capacities,
where goodput begins to level off. Additional results for these two
configurations (not plotted in panel (a)) show the hit rate falling from
0.84 to 0.55.

Panel (b) fixes SSD capacity at 2\,TB and uses admission limits of
8, 12, and 16.
It shows that GPU KV capacity can limit concurrency even when the
external tiers hold the working set. Raising the admission limit from 12
to 16 increases the mean number of running requests by only 0.8, but adds
2.9 waiting requests. Goodput, shown by the labels above the bars, falls
from 10.6 to 3.7 requests per minute despite the higher admission limit.
Increasing external cache capacity cannot resolve this GPU-space bottleneck.

These results identify two distinct capacity bottlenecks. External cache
capacity limits KV reuse: when the tiers become too small to retain KV
until reuse, evictions force recomputation instead of fast DRAM or SSD
loads and can trigger a goodput cliff. GPU KV capacity limits concurrency:
even with cache
hits, requests must wait when the GPU pool is full. Capacity should
therefore be added at the level that limits KV reuse or request concurrency.

\subsection{Where the compressible bits are}
\label{sec:bg-bits}
Fig.~\ref{fig:motiv}c summarizes all 48 layers of Qwen3-30B-A3B, each
with one key (K) tensor and one value (V) tensor. We split each BF16 value
into a high byte and a low byte. Within each tensor, we rank the 256 possible
values of each byte separately by frequency, then average the share of
occurrences in each rank group across the 96 tensors.
The high byte holds the sign bit and upper seven exponent bits. On average,
its seven most frequent values account for about 94\% of occurrences, and the top
14 account for 99.81\%, leaving 0.19\% for the remaining 242 values.
For the low byte, the top seven and top 14 values account for only 4.21\%
and 8.21\%, respectively. We observe a similar pattern in the other models
we evaluated. This concentration makes the high byte the
main target for lossless compression to increase effective KV cache capacity.

\subsection{Why not existing GPU compression methods}
\textbf{General-purpose compression.} Libraries such as
nvCOMP~\cite{nvcomp} offer GPU compression and decompression of data blocks
across applications. These compressors are not specifically designed for
KV caches. We compare nvCOMP's ANS, Bitcomp and LZ4 implementations
with \sys{} on the same KV data and GPU (Table~\ref{tab:codec}). They save
less space and restore paged KV more slowly; nvCOMP ANS stores 72.5\% of
raw bytes and restores at 182\,GB/s, versus 70.9\% and 1182\,GB/s for \sys{}.

\textbf{Offline weight compression.} DFloat11, ZipNN and
ZipServ~\cite{dfloat11,zipnn,zipserv} compress static model weights offline
to reduce their storage footprint. ZipServ also fuses decompression with
matrix multiplication (GEMM). KV is generated online and consumed from
paged caches by attention, so these weight-oriented workflows and kernels
cannot be used directly.

\textbf{Transfer-oriented KV compression.} SplitZip~\cite{splitzip}
reduces communication between prefill and decode workers by compressing
KV for transfer. It decompresses into a contiguous buffer at 1215\,GB/s.
It then needs a separate pass to scatter the KV into the engine's cache pages.
On the same KV data and GPU, SplitZip stores 76.8\% of the raw bytes and
restores paged KV at 671\,GB/s. \sys{} stores 70.9\% and restores KV
directly into the pages at 1182\,GB/s (Table~\ref{tab:codec}).

Compression must also reduce allocated cache space; shorter outputs
alone do not increase usable capacity.

%% file: sections/03-overview.tex
\section{System Overview}
\label{sec:overview}
\begin{figure*}[t]
\centering
\includegraphics{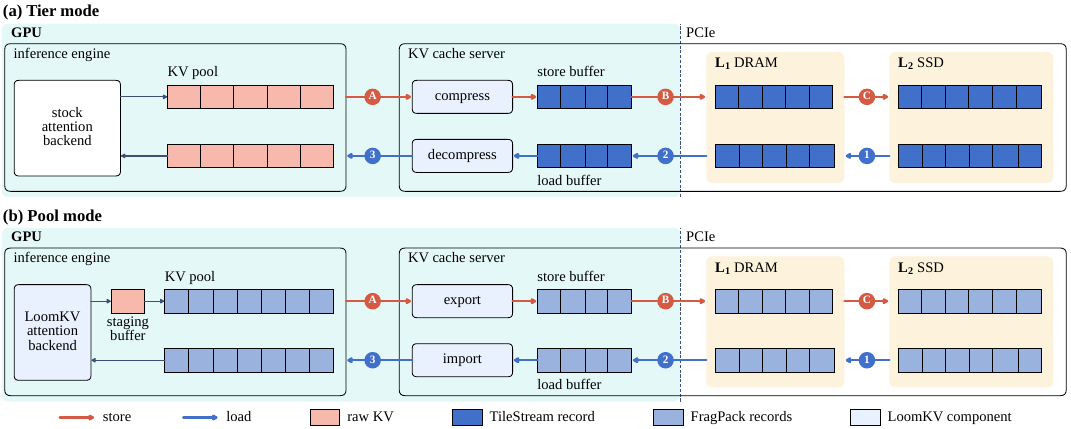}
\caption{\sys{} in an inference engine and a KV cache server. (a) Tier mode: compression in the tiers. (b) Pool mode: compression in the GPU pool.}
\label{fig:arch}
\end{figure*}

\textbf{Two modes.} \sys{} supports two modes sharing the lossless encoding
scheme of \S\ref{sec:codec}. In \emph{tier mode} (Fig.~\ref{fig:arch}a), the
inference engine keeps uncompressed KV in its GPU pool and uses its
existing attention backend. The KV cache server compresses KV for
storage in DRAM or SSD and restores its original bits when loading it
into the GPU pool. This increases external cache capacity while preserving
model execution and the cache server's tier management (\S\ref{sec:lmc}).

In \emph{pool mode} (Fig.~\ref{fig:arch}b), the inference engine keeps
its GPU KV pool compressed and reads it through a backend that fuses
decompression with decode attention (\S\ref{sec:hbm}), increasing GPU KV
capacity. Compression preserves every bit of the stored KV. The fused
attention kernel uses a different summation order from the original backend,
so its outputs may differ slightly due to floating-point rounding
(\S\ref{sec:eval-cost}).

\textbf{KV records.} In Fig.~\ref{fig:arch}, a \emph{record} is the unit of
compressed KV. Its granularity and layout depend on how the GPU decompresses
it. In tier mode, the cache server loads a whole \emph{chunk} into GPU
global memory on a hit. One \emph{TileStream record} therefore contains
both K and V for a block of 128 tokens across all layers.

In pool mode, each chunk has two \emph{FragPack records} per KV head in
each layer, one for K and one for V. The attention kernel uses the K record
to compute attention weights and the V record to accumulate the output.
This granularity lets a thread block stage
whole compressed records in the limited shared memory of a streaming
multiprocessor (SM). FragPack orders the data by the threads that consume
them, so each thread can decompress its operands directly into registers
for attention.

\textbf{Data paths (Fig.~\ref{fig:arch}).} The cache server is a separate process that reaches the engine's blocks through CUDA IPC, so its kernels read and write them directly. In tier mode, the engine's connector asks the server to store a chunk after the step that completes it. The server compresses the chunk into a GPU buffer~\pathstep{A}, copies the TileStream record into $L_1$~\pathstep{B} and also writes it to $L_2$~\pathstep{C}, because the server writes every object to both tiers (\emph{write-through}). On a lookup, the server returns the longest prefix it holds and prefetches chunks held only in $L_2$ into $L_1$~\pathstep{1}. A load copies the records into GPU buffers~\pathstep{2} and decompresses them into the engine's blocks~\pathstep{3}.

In pool mode, the K and V of new tokens are first kept raw in a staging buffer. When a block is complete, the attention backend compresses it into pages of the pool, and attention reads these pages directly. A store and a load take the same steps as in tier mode, but the cache server compresses nothing. It groups the FragPack records of a complete block into one cache object in a GPU buffer~\pathstep{A} and, on a load, imports them into the pool's pages~\pathstep{3}, so the tiers hold the pool's records unchanged. In both modes, only compressed bytes cross PCIe.

The next four sections describe the code and its allocation (\S\ref{sec:codec}),
TileStream and its decompressor (\S\ref{sec:lmc}), FragPack and fused attention
(\S\ref{sec:hbm}), and transfers without synchronization stalls (\S\ref{sec:sync}).

%% file: sections/04-codec.tex
\section{A Lossless KV Code}
\label{sec:codec}
Both modes of \sys{} share a lossless \emph{code} that stores each BF16
scalar's low byte verbatim and replaces its high byte with a fixed-width
\emph{symbol} indexing a static \emph{codebook}.

\subsection{Symbols, codebooks, and calibration}
\label{sec:code-sym}
\begin{figure}[t]
\centering
\includegraphics{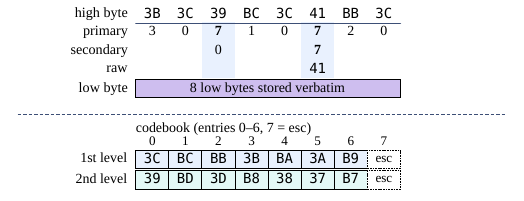}
\caption{The lossless KV code on eight BF16 scalars. Low bytes stay
verbatim; high bytes are coded through a calibrated two-level 3-bit
codebook per KV head in each layer's K or V tensor (entry~7 marks an escape).}
\label{fig:bits}
\end{figure}

\textbf{Byte-aligned, fixed-width symbols.} Because the BF16 low byte carries
7.97 bits of entropy out of 8, \sys{} stores it verbatim. For the high byte,
\sys{} codes all eight bits together rather than isolating the 8-bit exponent
as SplitZip~\cite{splitzip} does: because the exponent straddles the high and
low bytes, coding only the exponent shrinks records by only about 0.3\% while
breaking byte alignment and adding bit-shift and reassembly instructions on
every scalar.

\sys{} also avoids variable-length entropy codes such as Huffman coding and
asymmetric numeral systems (ANS). Because a GPU \emph{warp} executes 32 \emph{lanes} in lockstep, fast
decompression requires every lane to read its symbols at fixed offsets and
execute identical lookup instructions. Under a variable-length code, each
symbol's bit offset depends on all preceding symbols; even interleaved
multi-stream variants~\cite{giesen} still decompress several times slower on
GPUs than fixed-width lookups.

\textbf{Two-level 3-bit code.} Each high byte receives a 3-bit \emph{primary
symbol}: values 0--6 index the seven most frequent high bytes in the
codebook's first row, and value~7 is an \emph{escape}. We choose three bits
because wider symbols (such as 4~bits) would spend an extra bit on every
scalar even though the top seven values already cover roughly 94\% of high
bytes, and eight 1-byte entries fit the 8-byte register input of a single GPU
byte-permute instruction (\texttt{PRMT}). Because about 6\% of high bytes
escape the first row, writing an 8-bit raw byte on every primary escape would
add $0.06 \times 8 = 0.48$ bits per scalar; each primary escape instead emits
a 3-bit \emph{secondary symbol} indexing the next seven frequent values in the
codebook's second row. Only when a high byte falls outside all 14 entries does
secondary symbol~7 escape to a verbatim 8-bit raw byte (0.19\% of high bytes).
Fig.~\ref{fig:bits} traces this code on eight BF16 scalars: low bytes stay
verbatim, common high bytes (\texttt{3C}, \texttt{3B}) map directly to primary
symbols~0 and~3, \texttt{39} escapes once to secondary symbol~0, and rare
\texttt{41} escapes twice to a raw byte. Pool mode widens the second level to
4~bits (15 secondary entries) to keep fused decompression loop-free
(\S\ref{sec:fp-layout}).

\textbf{Per-head codebooks and calibration.} Sharing one global codebook across
all KV loses compression because each layer's K and V tensors have different
exponent ranges, whereas per-channel codebooks add negligible entropy gain
while forcing lanes to switch codebooks across channels. Because adjacent
scalars handled by a lane belong to the same KV head within one layer's K or V
tensor, \sys{} keeps one codebook per KV head in each layer's K or V
tensor—the finest granularity
at which each lane consults a single codebook. Codebooks remain static after
server startup: \sys{} provides a generic fallback codebook and calibrates
model-specific codebooks offline from a single 1024-token forward pass,
storing a codebook digest in record headers to prevent mismatches. One offline
calibration generalizes across prompt domains and sequence lengths.

\textbf{Extension to FP8 KV.} For 8-bit FP8 (E4M3) KV, the same two-level code
stores the three mantissa bits verbatim and codes the upper five bits (sign
and exponent) with FP8 codebooks. Because FP8 channel magnitudes vary within a
head, \sys{} subtracts a static per-channel integer exponent offset before
compression and adds it back during decompression so all channels within a
head share one codebook. Since rotary position embeddings (RoPE) rotate K
channel pairs with token position, offline calibration re-rotates captured K
tensors across the model's 256K context before fitting K's offsets; BF16 KV
needs no per-channel offsets and thus no RoPE re-rotation.

\subsection{From saved bytes to capacity}
\label{sec:code-cap}
Compressed records vary in byte length with their escape counts, whereas
external cache tiers and GPU KV pools are provisioned around fixed-size units.
Each mode converts saved bytes into capacity to match its allocator.

\textbf{Exact-length tier objects.} In tier mode, the KV cache server treats
the raw chunk size only as an upper bound: once GPU compression yields a
record's actual byte length, the server allocates an $L_1$ DRAM object of that
exact length and writes it unchanged to $L_2$ SSD. On $L_2$ write completion,
the server records the compressed length in host metadata so later prefetches
from $L_2$ reserve exact-sized $L_1$ buffers before issuing SSD reads. Neither
the variable-size $L_1$ allocator nor byte-based LRU eviction requires
changes.

\textbf{Smaller fixed-size GPU pages.} Pool mode preserves the engine's
fixed-size page-table layout so GPU memory management stays unchanged, but
shrinks each physical page to cover the typical compressed length of a block's
records, yielding proportionally more KV blocks from the same GPU budget
(\S\ref{sec:fp-page}).

%% file: sections/05-lmcache.tex
\section{Tier Mode: TileStream and Parallel Decompression}
\label{sec:lmc}
In tier mode, the KV cache server compresses each chunk into a
\emph{TileStream record} and decompresses fetched records straight into the
inference engine's scattered GPU pages.

\subsection{The TileStream layout}
\label{sec:ts-layout}
\begin{figure}[t]
\centering
\includegraphics{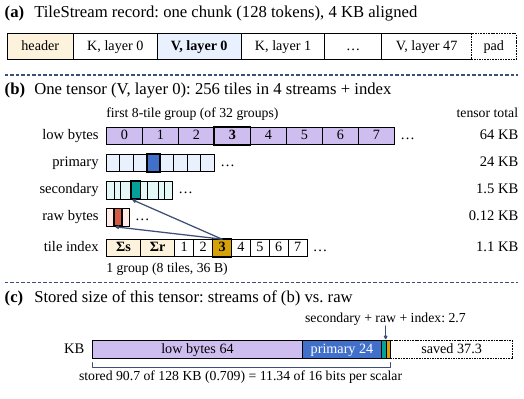}
\caption{The TileStream layout (mean full-model sizes for Qwen3-30B-A3B).
(a)~One record spanning a 128-token chunk across 48 layers (96 K/V tensors),
4\,KB aligned. (b)~One 4-head K/V tensor (256 tiles; 128 tiles per rank at
TP~2): first 8-tile index group with tile~3 highlighted, and full-tensor byte
totals on the right. (c)~The same tensor's stored size: the totals of (b)
summed against its 128\,KB raw size.}
\label{fig:format}
\end{figure}

\textbf{Whole record (Fig.~\ref{fig:format}a).} On Qwen3-30B-A3B, one
TileStream record holds 96 K/V tensors (one K and one V per layer across 48
layers) behind a 4\,KB-aligned header of per-tensor offsets (V, layer~0
expanded in panel~b), followed by 4\,KB
padding for \texttt{O\_DIRECT} SSD transfers. Incompressible chunks fall back
to raw storage so worst-case size never exceeds raw KV. Kernels address engine
blocks through strides to accommodate different engine K/V layouts and execute
independently per tensor-parallel (TP) rank (12\,MB raw full-model chunk on
Qwen3-30B-A3B, or 6\,MB per rank at TP~2).

\textbf{Streams and grouped tile index (Fig.~\ref{fig:format}b).} Each K/V
tensor groups bytes by type into four contiguous streams—low bytes, primary
symbols, secondary symbols, and raw bytes—so GPU threads read fixed-width data
via coalesced loads while isolating variable-length escapes to the last two
streams. Each tensor splits into 256-scalar tiles (256 tiles per 4-head tensor
on Qwen3-30B-A3B, or 128 per rank at TP~2): one GPU lane reads or writes eight
BF16 scalars in a single 128-bit vector instruction (16\,B; 64-bit for FP8), so
a 32-lane warp naturally covers $32 \times 8 = 256$ scalars as one tile. While
a tile's low-byte and primary offsets are fixed, its secondary and raw offsets
depend on all prior escapes. To let warps decompress tiles in parallel without
scanning earlier tiles, each K/V tensor appends a grouped \emph{tile index}.
Storing two 32-bit prefix counts per tile would cost 0.25 bits/scalar;
grouping eight tiles (32 groups per tensor) halves this to 0.14 bits/scalar
(36\,B/group): two 32-bit base counts ($\Sigma_s, \Sigma_r$) for tile~0 plus
seven 32-bit words packing 16-bit group-relative secondary and raw offsets for
tiles~1--7. Fig.~\ref{fig:format}b shows the first 8-tile group alongside
full-tensor totals (64, 24, 1.5, 0.12, and 1.1\,KB): to reach tile~3, a warp
computes its fixed low/primary positions and adds word~3 to ($\Sigma_s,
\Sigma_r$) to jump straight to tile~3's secondary and raw slices (arrows).

\textbf{Stored size and bit budget (Fig.~\ref{fig:format}c).} The streams and the index of
one tensor add up to 90.7\,KB of its 128\,KB raw size. Per scalar, the low-byte
and primary streams cost a fixed 8 and 3 bits, and the secondary stream, the
raw stream, and the index add $3p_e + 8p_r + 0.14$ bits. At Qwen3-30B-A3B's measured escape rates
($p_e = 0.061$, $p_r = 0.0019$), escapes and the index add only 0.34 bits,
totaling 11.34 of 16 bits (ratio 0.709, saving 4.66 bits) with under 1\%
metadata and alignment overhead.

\subsection{Parallel decompressor}
\label{sec:ts-dec}
\begin{figure*}[t]
\centering
\includegraphics{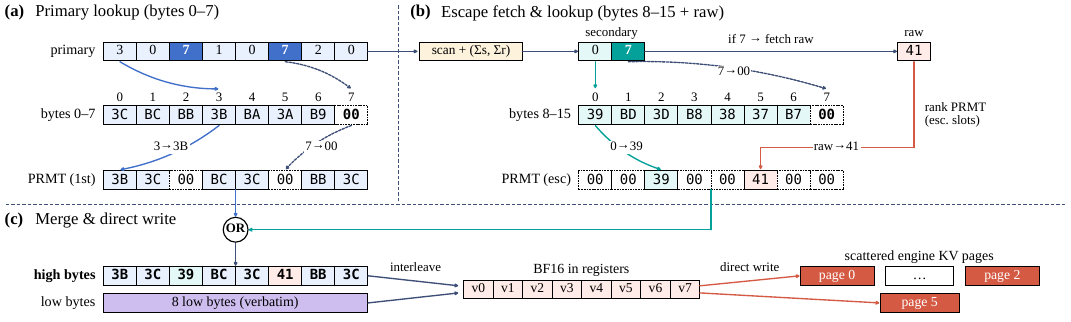}
\caption{TileStream decompression with explicit branchless \texttt{PRMT} lookup
for one lane (eight BF16 scalars) inside a 32-lane warp (one 256-scalar tile).
(a)~Primary lookup in codebook bytes~0--7 (escape slot~7 holds \texttt{00}).
(b)~Warp scan and escape lookup in bytes~8--15 plus raw fetch, placed into
escaped slots via rank \texttt{PRMT}. (c)~Bitwise OR of both \texttt{PRMT}
outputs, interleaving with verbatim low bytes, and direct register writes to
scattered engine KV pages.}
\label{fig:tsdec}
\end{figure*}

During a load, each warp decompresses one 256-scalar tile directly into the
engine's scattered KV pages, with each of its 32 lanes handling eight adjacent
scalars from one KV head. Fig.~\ref{fig:tsdec} traces one lane through four
steps (with branchless lookup):

\textbf{1. Mark escapes.} The lane loads its eight 3-bit primary symbols as
one 24-bit word and unpacks them into eight 4-bit nibbles (matching
\texttt{PRMT}'s 4-bit selector width). Bitwise-ANDing the three value bits of
each nibble flags every primary escape (symbol~7 at positions~2 and~5) and
counts escapes locally without cross-lane communication.

\textbf{2. Warp scan.} Two warp-level exclusive prefix sums over the lanes'
primary- and secondary-escape counts, added to the tile's index offsets
($\Sigma_s, \Sigma_r$), give each lane its exact positions in the secondary
and raw streams (Fig.~\ref{fig:tsdec}b), from which it loads up to one 24-bit
secondary word and any needed raw bytes.

\textbf{3. Branchless PRMT lookup.} Each head's codebook is repacked into
16\,B in registers: primary values in bytes~0--6, secondary values in
bytes~8--14, and \texttt{00} in escape slots~7 and~15. A first \texttt{PRMT}
lookup resolves all non-escaped primary symbols while writing \texttt{00} at
escaped slots (Fig.~\ref{fig:tsdec}a); a second \texttt{PRMT} pass resolves
secondary symbols, merges any raw bytes, and uses a rank \texttt{PRMT} to
place them into the escaped slots with \texttt{00} elsewhere
(Fig.~\ref{fig:tsdec}b). A single bitwise OR produces all eight high bytes
with zero branch divergence.

\textbf{4. Reassemble and direct write.} The lane loads its eight verbatim low
bytes, interleaves them with the decompressed high bytes in registers (for
FP8, it adds per-channel offsets and reattaches the 3-bit mantissa), and
writes the eight scalars straight into the engine's target KV page
(Fig.~\ref{fig:tsdec}c), avoiding any staging buffer or scatter pass. To guard
against corrupt records, write addresses come exclusively from host-validated
metadata and all record-derived read offsets are clamped to the load buffer.

\subsection{Direct-from-pages compressor}
\label{sec:ts-enc}
On the store path, three consecutive GPU kernels compress a batch of chunks
with no host synchronization between them: a \emph{count} kernel reads KV
directly from the engine's paged blocks and tallies per-tile escapes, a
single-cooperative-thread-array (CTA) \emph{layout} kernel computes stream offsets and flags
incompressible chunks for raw fallback, and a \emph{compact} kernel writes the
aligned header, streams, and tile index into the store buffer. Reading
directly from paged blocks instead of first gathering chunks into a contiguous
buffer saves two full passes over raw KV and roughly two kernel launches per
chunk, reducing both memory traffic and GPU time-slice contention with the
inference engine.

%% file: sections/06-gpu-pool.tex
\section{Pool Mode: FragPack and Fused Decode Attention}
\label{sec:hbm}
In pool mode, the inference engine stores compressed \emph{FragPack records}
directly in its GPU KV pool, and an out-of-tree attention backend plugin
computes decode attention over them without engine source changes.

\subsection{FragPack records and compressed pages}
\label{sec:fp-layout}
\begin{figure*}[t]
\centering
\includegraphics{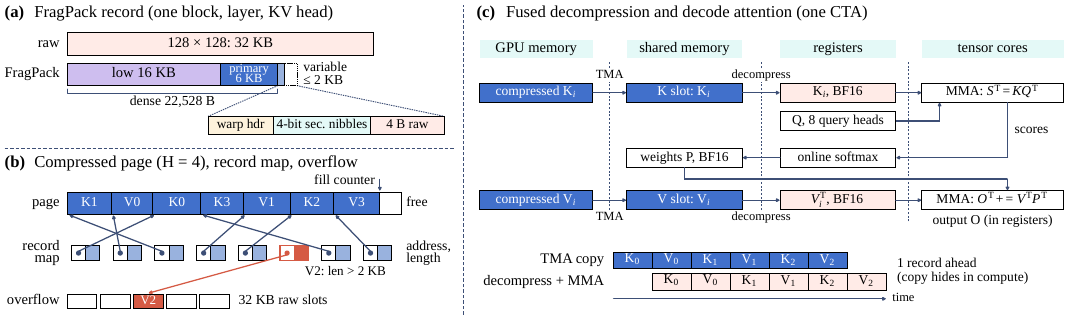}
\caption{Compressed GPU pool and fused decode attention in pool mode.
(a)~One FragPack record (K or V of a block, layer, and KV head) versus its
32\,KB raw size, with the three pieces of its variable part expanded.
(b)~One compressed page ($H = 4$, 8 records) filled in parallel completion
order up to the atomic fill counter, indexed by the per-layer record map;
\texttt{V2} overflows to a 32\,KB raw slot (length $> 2$\,KB).
(c)~Fused decompression and decode attention inside one CTA across GPU memory,
shared memory, registers, and tensor cores, with the record-ahead Tensor Memory
Accelerator (TMA)/compute timeline at the bottom.}
\label{fig:fppage}
\end{figure*}

\textbf{Granularity and MMA fragment order.} Each \emph{FragPack record} holds
$128\times 128$ scalars (32\,KB raw; Fig.~\ref{fig:fppage}a) for one
128-token block, layer, KV head, and tensor (K or V). The fused kernel runs
256-thread (8-warp) CTAs, each over a run of consecutive pages of one request
and KV head, and feeds every record it reads to tensor-core matrix
multiply-accumulate (MMA) instructions, where each thread supplies a fixed
64-scalar \emph{fragment} of the record from registers. FragPack
permutes scalars at compression time into this per-thread fragment order (K in
row order, V in transposed order $V^\top$), so every thread fetches its bytes
via wide, coalesced loads indexed solely by thread ID.

\textbf{Two-part layout and 4-bit secondary level.} Each record pairs a
fixed-length \emph{dense part}—16\,384\,B of low bytes plus 6144\,B of 3-bit
primary symbols (22\,528\,B)—with a \emph{variable part} capped at 2\,KB so
escapes stage in a fixed shared-memory buffer (records exceeding this cap fall
back to raw). Because each of the CTA's eight warps always decompresses its
own fixed 2048 scalars, FragPack needs no tile index; an eight-entry per-warp
prefix header suffices to locate every warp's secondary
and raw slices. As expanded in Fig.~\ref{fig:fppage}a, the variable part holds
this header, 4-bit secondary nibbles (15 codebook entries plus escape~15), and
4-byte raw entries (each storing a raw high byte and its escape position). Widening the second
level to 4~bits (\texttt{3+4}) keeps fused decompression loop-free: each of a
warp's 32 lanes patches at most one raw scalar (capping warps at 32 raw
scalars before raw fallback). On captured Qwen3-30B-A3B KV, 4-bit nibbles cut
raw scalars from 25 to 0.3 per record (max $\le 10$ per warp, 0\% cap-induced
fallback vs.\ 0.2\% with 3~bits) at +0.1\% record size; the variable part
averages 550\,B and never exceeds 1872\,B.

\textbf{Shared-budget pages, overflow, and record map.}\label{sec:fp-page}
Each compressed GPU page holds $2H$ records for one block and layer ($H = 4
\Rightarrow 8$ records in Fig.~\ref{fig:fppage}b). Rather than reserving the
worst-case 2\,KB per record, the $2H$ records share a pooled variable budget
of 1280\,B per record (above Qwen3-30B-A3B's fullest page at 1140\,B/record
and well above the 395--562\,B means across nine grouped-query attention (GQA)
models). For $H = 4$, a
page shrinks from 262\,144\,B to 190\,464\,B, yielding $1.38\times$ as many
blocks before deducting backend working buffers.

When a block finishes, its $2H$ records compress in parallel and claim space
in arbitrary completion order (\texttt{K1, V0, K0, K3, V1, K2, V3} in
Fig.~\ref{fig:fppage}b) by advancing a per-page \emph{fill counter} via
atomic compare-and-swap (CAS); if a record exceeds the remaining space, it
aborts without advancing the counter so shorter records can still fit. Records that overflow the page or exceed the
compression limits are stored raw (32\,KB) in a shared \emph{overflow area}
(\texttt{V2} in Fig.~\ref{fig:fppage}b); in Qwen3-30B-A3B serving runs, only
0.03\% of records fall back to raw (0.009\% from full pages). A per-layer \emph{record map} stores each record's (address, length),
pointing directly into either its page or an overflow slot. Storing the
variable-part length for compressed records and 32\,KB for raw records lets
attention detect raw fallback via length~$> 2$\,KB and read every
record directly without computing page-internal offsets.

\subsection{Fused decompression and decode attention}
\label{sec:fp-attn}
Because each decode step emits one token per request while reading the entire
context's KV, decompressing into a workspace first would write and re-read all
raw bytes every step. \sys{} instead decompresses each FragPack record
straight into registers for tensor-core MMAs, reading $\approx 0.70$ of raw KV
bytes and never writing decompressed KV back to GPU memory.

\textbf{Three-stage CTA dataflow and pipelining (Fig.~\ref{fig:fppage}c).}
Inside one CTA, each page's K and V records flow across GPU memory, shared
memory, registers, and tensor cores in three stages:
(1)~\emph{Aligned TMA staging:} the Tensor Memory Accelerator (TMA) copies
each record's exact byte length asynchronously into a dedicated shared-memory
K or V slot. Because TMA penalizes unaligned 128-byte
transfers, matching the slot's address remainder modulo~128 to the source
address avoids a $\approx 30\%$ bandwidth penalty.
(2)~\emph{Register decompression:} each thread decompresses its 64 scalars
into registers via \texttt{PRMT} primary lookup, warp-scan escape lookup, and
loop-free 4-bit secondary/raw patching (skipping decompression on raw records
for bitwise identical output), yielding $K_i$ and transposed $V_i^\top$
directly in MMA fragment order.
(3)~\emph{Tensor-core MMAs and online softmax:} using \texttt{m16n8k16} MMAs
($C \mathrel{+}= AB$), the K step multiplies $K_i$ (in $A$) by the eight GQA
query heads (in $B$) to obtain scores $S^\top = K Q^\top$ and online-softmax
BF16 weights $P$. Because warps partition 16 tokens in the K step but 16
output dimensions across all 128 tokens in the V step, the CTA exchanges $P$
through shared memory before accumulating
$O^\top \mathrel{+}= V^\top P^\top$ in registers.

Dual K/V shared-memory slots keep TMA one record ahead of computation across
the CTA's pages (\texttt{K0, V0, K1, V1, K2, V2}, bottom of
Fig.~\ref{fig:fppage}c): TMA fetches \texttt{V0} while threads process
\texttt{K0} and fetches \texttt{K1} once \texttt{K0}'s slot retires, hiding
copy latency behind decompression and MMAs.

\textbf{Prefill.} Because prefill is compute-dominated rather than KV-read
dominated, pool mode leaves prefill unfused: it decompresses cached pages
into a temporary BF16 workspace and runs the engine's stock attention kernel,
adding at most 3\% to prefill time.

\subsection{Open blocks and CUDA graphs}
\label{sec:fp-state}
\textbf{Open blocks.} Only complete 128-token blocks are compressed; each
running request keeps its in-progress \emph{open block} raw in a per-request
\emph{staging buffer} that attention reads alongside the compressed pages.
Before each step, a lightweight host hook reads scheduler metadata to update
request block lists, recycle finished staging buffers, assign buffers to newly
opened blocks, and populate a fixed-capacity list of blocks hitting token~128.
During the step, each layer compresses those newly completed blocks right
after attention (one CTA per record).

\textbf{CUDA graph compatibility.} Serving engines replay full decode steps as
single CUDA graphs, which requires kernel grids and argument buffers to stay
fixed for a given padded batch shape. Because per-step attention scheduling
computes CTA work assignments on the GPU and block compression reads the
fixed-capacity list populated outside the graph, all per-layer launch
parameters depend solely on padded batch shape, preserving full CUDA-graph
capture and replay.

%% file: sections/06s-sync.tex
\section{Avoiding Synchronization Stalls in KV Transfers}
\label{sec:sync}
\begin{figure*}[t]
\centering
\includegraphics{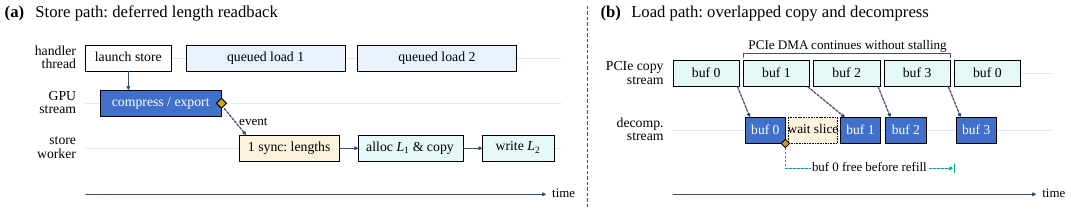}
\caption{Asynchronous KV transfers in the cache server. (a)~Store path: the
handler launches compression or export and immediately serves queued loads
while a store worker waits on a CUDA event, performs one batched length
readback, allocates and copies $L_1$ objects, and writes through to $L_2$.
(b)~Load path: rotating four GPU load buffers decouples the PCIe copy stream
from the GPU decompress/import stream via per-buffer events, hiding
multi-process GPU time-slice waits behind PCIe transfers.}
\label{fig:transfer}
\end{figure*}

The KV cache server executes all stores and loads for a GPU on one handler
thread. Variable-length compression creates two stall hazards: reading
GPU-computed compressed lengths back to the host would block queued loads on
the handler thread, and multi-process GPU time-slice waits during
decompression would stall PCIe transfers if copies and decompression shared a
single buffer. Fig.~\ref{fig:transfer} shows how \sys{} eliminates both
stalls.

\subsection{Store path: deferred length readback}
\label{sec:sync-store}
To move length readback off the handler thread, \sys{} splits stores across
two threads (Fig.~\ref{fig:transfer}a). For each batch of uncached chunks (up
to 32), the handler claims a GPU \emph{store buffer}, launches the
three-kernel compressor (\S\ref{sec:ts-enc}), records a CUDA event, hands the batch to a background
\emph{store worker}, and immediately returns to serve queued loads. On its own
CPU thread and CUDA stream, the store worker waits on the event, reads all
batch lengths in one host synchronization, allocates exact-length $L_1$
objects, copies the records into $L_1$, and triggers the server's
write-through to $L_2$. If all store buffers fill under bursts, the handler
applies backpressure until one frees rather than dropping stores.

\subsection{Load path: overlapped copies and decompression}
\label{sec:sync-load}\label{sec:ts-load}
Because host metadata already holds each pinned $L_1$ object's exact length,
loads need no length readback and run directly on the handler thread, skipping
chunks already present in the engine's local prefix cache. To prevent
multi-process GPU compute time-slice waits from stalling PCIe DMA (which does
not wait for compute slices), \sys{} pipelines PCIe copies and GPU
decompression on two CUDA streams rotating through a ring of GPU \emph{load
buffers} (four in our deployment; Fig.~\ref{fig:transfer}b). Each
decompression waits via a CUDA event only on its own buffer's copy, and each
buffer refill waits only on that buffer's prior decompression, hiding
time-slice pauses behind continuous PCIe transfers. Kernel launch descriptors are cached once per batch shape. When a
batch includes an incompressible raw chunk, or when its first chunk lies
partly inside the engine's local prefix-cache hit so only the remaining tokens
may be written, \sys{} falls back to decompressing into a scratch buffer and
scattering via the server's stock transfer kernel.

\subsection{Extending both paths to pool mode}
\label{sec:sync-blob}
Pool mode reuses both asynchronous pipelines with three localized changes:
(1)~\emph{Object format:} each cache object bundles all layers' FragPack
records for one complete 128-token block behind a 4\,KB-aligned header (KV
shape, codebook digest, record lengths), keeping one uniform representation
across GPU, $L_1$, and $L_2$ (requiring chunk size = block size = 128).
(2)~\emph{Store path:} an \emph{export} kernel replaces compression,
gathering records from pool pages or overflow into a store buffer; deferred
length readback, $L_1$ copy, and $L_2$ write-through stay unchanged.
(3)~\emph{Load path:} an \emph{import} kernel replaces decompression,
validating headers, packing records into pool pages via atomic fill counters
(\S\ref{sec:fp-page}), and publishing record-map entries. Because corrupt headers or a full overflow
area can fail an import, the handler performs one status readback after
queueing all imports for the request—avoiding mid-pipeline stalls—and the
engine recomputes any failed batch.

%% file: sections/08-evaluation.tex
\section{Evaluation}
\label{sec:eval}
We evaluate \sys{} around six questions, one per subsection:

\textbf{Q1}: how much space does the code save across models and workloads?

\textbf{Q2}: are its compression, decompression, and fused-attention kernels fast
enough to stay off the critical path?

\textbf{Q3}: how much does tier mode improve goodput and TTFT when external
cache tiers are short of space?

\textbf{Q4}: how much does pool mode improve them when the GPU KV pool limits
concurrency?

\textbf{Q5}: do both modes also help on a second production-derived workload?

\textbf{Q6}: what overheads does compression introduce, and which outputs
remain exact?

\subsection{Setup}
This subsection describes the shared serving testbed used across our main
AgentX experiments; setup details specific to the code microbenchmarks,
kernel benchmarks, and the second workload appear in Q1, Q2, and Q5.

\textbf{Hardware and models.} The server has 1\,TB of DRAM, 15\,TB of SSD,
and two H800 GPUs (80\,GB, 700\,W). Unless noted, we serve
Qwen3-30B-A3B-Instruct at TP\,2 (mixture-of-experts, MoE; 96\,KB of BF16 KV per token), alongside
GLM-4-9B-Chat-1M (dense, 80\,KB/token). Q1 additionally evaluates stored KV
ratios across nine dense and MoE models from 4B to 72B parameters. The
inference engine is unmodified vLLM~0.30 with 128-token blocks and FA3 as its
stock attention backend.

\textbf{Cache server and baselines.} Every serving configuration runs
LMCache~0.5.6 with 128-token chunks, bounded DRAM ($L_1$) and SSD ($L_2$)
tiers, and LRU eviction. The raw BF16 baseline runs unmodified LMCache; the
lossy baseline configures vLLM with uncompressed FP8 KV (halving KV size in
both the GPU pool and external tiers). In Q2, we additionally benchmark
\sys{}'s kernels against SplitZip, ExANS, and three nvCOMP compressors (ANS,
Bitcomp, LZ4), and compare offline sizes with ZipServ and ZipFusion.

\textbf{Main workload.} Our primary serving workload is
AgentX~\cite{agentx}, replaying the 393 Claude Code sessions of
cc-traces~\cite{cctraces} as session trees with subagents and recorded think
times; all reported metrics cover only the profiling window after warm-up.
Sessions share no prefix, and prompts reach 254K tokens (median 89K). Each GPU
reserves 54\,GB for KV (1.18\,M raw tokens); in pool mode, the staging
buffers, overflow area, and record map are deducted from this same 54\,GB
budget.

\textbf{Admission and metrics.} Because the harness keeps one request in
flight per live session, a FIFO proxy in front of vLLM admits at most $S$
requests concurrently so engine queueing does not mask cache effects. We
default to $S = 8$, which keeps the raw GPU pool free of internal queueing
(Q4 sweeps $S$ to 16). We report \emph{engine TTFT} (harness TTFT minus proxy
wait), \emph{goodput} (requests/min meeting engine TTFT $\le 2$\,s and
inter-token latency (ITL) $\le 40$\,ms), \emph{hit rate} (share of prompt tokens
served from cache), and \emph{throughput} (prompt tokens/s).

\textbf{Sizing and repeats.} With $S$ fixed, live sessions $P$ set the reused
working set $W$ (306\,GB raw at $P = 32$; 641\,GB at $P = 64$). Most
experiments use $P = 64$, 400\,GB DRAM, and SSD caps from 400\,GB to 2\,TB
over an 1800\,s profiling window ($\approx 600$ requests).
We run every experiment three times and report the median (each end-to-end
table row is the median-goodput run). Where tiers hold the working set,
repeats vary by roughly 1\% in throughput and 4\% in goodput; near a capacity
cliff, we also report the full repeat range.

\subsection{Q1: Compression ratio}
\label{sec:eval-codec}
\begin{table}[t]
\centering
\caption{Stored Ratio of KV by Model}
\label{tab:models}
\setlength{\tabcolsep}{3pt}
\begin{tabular}{lrrrrrrr}
\toprule
 & & \multicolumn{4}{c}{TileStream} & \multicolumn{2}{c}{FragPack} \\
\cmidrule(lr){3-6}\cmidrule(lr){7-8}
Model & $L{\times}H$ & generic & calib. & bound & FP8 & ratio & var.\,B \\
\midrule
Qwen3-4B & 36$\times$8 & 0.729 & 0.709 & 0.676 & 0.873 & 0.705 & 562 \\
Qwen3-8B & 36$\times$8 & 0.732 & 0.709 & 0.676 & 0.872 & 0.705 & 561 \\
Qwen3-32B & 64$\times$8 & 0.731 & 0.708 & 0.674 & 0.873 & 0.703 & 515 \\
Qwen3-30B-A3B & 48$\times$4 & 0.764 & 0.709 & 0.676 & 0.854 & 0.704 & 550 \\
Qwen2.5-72B & 80$\times$8 & 0.722 & 0.707 & 0.672 & 0.866 & 0.702 & 484 \\
Llama-3-8B & 32$\times$8 & 0.732 & 0.707 & 0.672 & 0.861 & 0.702 & 479 \\
Llama-3.1-8B & 32$\times$8 & 0.730 & 0.706 & 0.672 & 0.861 & 0.702 & 464 \\
Llama-3.1-70B & 80$\times$8 & 0.740 & 0.705 & 0.669 & 0.853 & 0.700 & 395 \\
GLM-4-9B-1M & 40$\times$4 & 0.730 & 0.707 & 0.673 & 0.865 & 0.702 & 459 \\
\bottomrule
\end{tabular}\\[2pt]
\parbox{\linewidth}{\footnotesize 4096 tokens of an agent session; codebooks calibrated on the first 1024.
$L{\times}H$: layers $\times$ KV heads. TileStream records: BF16 with a generic or
calibrated codebook, their order-0 bound, and calibrated FP8. FragPack
records: BF16 ratio and mean variable part in bytes.}
\end{table}

Across all nine models, \sys{} stores BF16 KV at 0.70--0.71 of its raw size,
staying within 0.033--0.036 of the order-0 entropy bound.

\textbf{Cross-model compression ratio.} Table~\ref{tab:models} shows that
calibrated TileStream records compress BF16 KV tightly and uniformly across
all nine Qwen, Llama, and GLM models (0.705--0.709, within 0.033--0.036 of the
order-0 bound). On Qwen3-30B-A3B, this 0.709 ratio breaks down into 0.676
entropy, 0.024 fixed-width symbol overhead, and 0.009 for the grouped tile
index and 4\,KB-aligned header. Calibration improves every model over the
generic fallback (most visibly on Qwen3-30B-A3B, from 0.764 to 0.709), and
compresses FP8 KV to 0.853--0.873. FragPack reaches slightly tighter BF16
ratios (0.700--0.705) because its per-warp header is smaller than a tile
index, and its variable part stays well under the 1280\,B per-record page
budget across every model (\texttt{var.\,B} in Table~\ref{tab:models}); on
held-out FP8 agent KV up to 240K tokens, FragPack compresses to 0.838.

\textbf{Workload and context-length stability.} Calibrated codebooks depend
on the model rather than the prompt domain or sequence position. One offline
forward pass over 1024 tokens produces them, and they occupy 8.4\,KB on
Qwen3-30B-A3B (22\,B for each layer, K or V, and KV head; FP8 adds 48\,KB of
per-channel exponent offsets). Codebooks calibrated on one agent session store prose,
source code, Chinese chat, math, synthetic needles, and random tokens at
0.706--0.709 BF16—within 0.001 of domain-specific calibrations (and within
0.018 for FP8, at 0.852--0.871)—and stay flat at 0.709--0.710 across every
position of a 240K-token context.

\begin{figure}[t]
\centering
\includegraphics{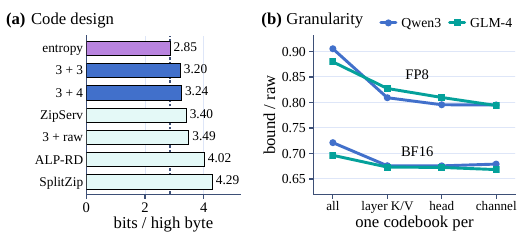}
\caption{Code design ablations on held-out agent KV (Qwen3-30B-A3B and
GLM-4-9B). (a)~Average coded bits per BF16 high byte across code designs.
(b)~Order-0 entropy bounds by codebook granularity for BF16 and FP8
(per-head FP8 includes per-channel exponent offsets).}
\label{fig:codec-abl}
\end{figure}

\textbf{Design choices (Fig.~\ref{fig:codec-abl}).} Panel~(a) explains why our
fixed-width code approaches the 2.85-bit order-0 entropy bound: two-level
\texttt{3+3} (3.20~bits/high byte) beats every alternative—6\% fewer bits than
ZipServ~\cite{zipserv}, 8\% fewer than single-level \texttt{3+raw}, 20\% fewer
than ALP-RD's positioned exceptions~\cite{alp}, and 25\% fewer than
SplitZip's 4-bit exponent code with 56-bit out-of-band
exceptions~\cite{splitzip}. Pool mode's \texttt{3+4} adds only 1.2\% to
high-byte bits (+0.1\% to total FragPack record size) while
enabling loop-free fused decompression. Panel~(b)
justifies per-head codebooks: moving from one global codebook to per-layer K/V
tensors captures almost all BF16 entropy gain across both models, per-head stays
flat, and per-channel
adds no benefit (slightly overfitting held-out Qwen3 KV). For FP8, subtracting
static per-channel exponent offsets lets per-head
codebooks match per-channel bounds on Qwen3 (0.795 vs.\ 0.795) and close
roughly half the remaining gap on GLM-4 (0.810 vs.\ 0.794).

\subsection{Q2: Kernels}
\label{sec:eval-kern}
\begin{table}[t]
\centering
\caption{Stored Ratio and Throughput (Raw-Equivalent {\upshape GB/s})}
\label{tab:codec}
\setlength{\tabcolsep}{2.4pt}
\begin{tabular}{lrrrr}
\toprule
Compressor & ratio & comp. & paged & contig. \\
\midrule
Raw copy & 1.000 & -- & 1452 & 1523 \\
\sys{} & 0.709 & \textbf{535} & \textbf{1182} & 1182$^{\mathrm{a}}$ \\
SplitZip~\cite{splitzip} & 0.768 & 328 & 671 & \textbf{1215} \\
ExANS~\cite{exans} & \textbf{0.698} & 105 & 187 & 187$^{\mathrm{a}}$ \\
nvCOMP ANS & 0.725 & 112 & 182 & 267 \\
nvCOMP Bitcomp & 0.876 & 114 & 152 & 207 \\
nvCOMP LZ4 & 1.000 & 9 & 150 & 206 \\
ZipServ code$^{\mathrm{b}}$ & 0.722 & -- & -- & -- \\
ZipFusion blocks$^{\mathrm{b}}$ & 0.79 & -- & -- & -- \\
\bottomrule
\end{tabular}\\[2pt]
\parbox{\linewidth}{\footnotesize One H800, Qwen3-30B-A3B KV (3.2\,GB, 256 chunks per call). Paged: decompress into a paged pool's blocks, with one scatter launch per call where the compressor cannot write them; contig.: into the compressor's own buffer. Bold: best compressor.
$^{\mathrm{a}}$Decompresses directly into scattered paged blocks with no separate scatter pass, so paged and contiguous throughputs are identical.
$^{\mathrm{b}}$Size only, computed offline: ZipServ's kernels~\cite{zipserv} decompress only weights (we add \sys{}'s tile index), ZipFusion's~\cite{zipfusion} have no tier path.}
\end{table}

\sys{}'s kernels keep compression and decompression off the critical path in
both modes: TileStream decompresses paged blocks at 81\% of raw copy speed and
hides behind PCIe copies under GPU contention, while FragPack's fused BF16
decode kernel outperforms both FA3 and FlashInfer on nearly every batch shape.

\textbf{Tier-mode compression and decompression.} Table~\ref{tab:codec}
benchmarks single-GPU throughput with byte-for-byte verification. On stores,
\sys{} compresses directly from paged blocks 1.6$\times$ faster than SplitZip
and 4.7--59$\times$ faster than ExANS and nvCOMP. On loads, writing straight
from registers into scattered KV pages achieves 81\% of
raw paged copy speed—an order of magnitude above PCIe bandwidth—avoiding the
extra scatter pass that nearly halves contiguous-only SplitZip's throughput
while storing 8\% fewer bytes. Variable-length ANS
decompressors (ExANS, nvCOMP ANS) run 6.3$\times$ slower despite ExANS's 1.6\%
tighter ratio; general-purpose Bitcomp saves little space; and LZ4 finds no
byte-level matches in interleaved BF16 (ratio 1.0005, stored raw). Offline size
comparisons with ZipServ and ZipFusion confirm the same compression margin
(Table~\ref{tab:codec}).

\textbf{Overlapped loads beside a busy engine.} In serving, LMCache shares
the GPU with vLLM as a separate process, so decompression can interfere with
inference~\cite{shadowserve}. Without CUDA Multi-Process Service (MPS), decompression kernels wait for
GPU compute time slices whereas PCIe DMA copies do not. Loading a 110K-token
prefix while another process saturates the GPU with BF16 matrix
multiplications exposes this stall: with one load buffer, every PCIe copy
waits for the prior decompression to get a time slice, inflating latency to
288--294\,ms (more than double raw copies at 117--141\,ms). Rotating four load
buffers cuts latency to
90\,ms—close to 83\,ms with one buffer on an idle GPU and matching PCIe
transfer time for 0.71 of the raw bytes—keeping decompression hidden behind
PCIe transfers.

\begin{figure}[t]
\centering
\includegraphics{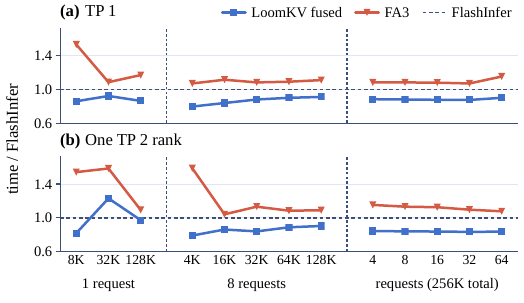}
\caption{Pool-mode decode attention time relative to FlashInfer (dashed line
at 1.0) at full GPU clock, across one-request, eight-request, and 256K-token
batch sweeps. (a)~TP~1 (4 KV heads). (b)~One TP~2 rank (2 KV heads).}
\label{fig:fpkern}
\end{figure}

\textbf{Pool-mode fused decode attention.} Fig.~\ref{fig:fpkern} compares our
fused BF16 decode kernel against FA3 (as configured by vLLM; FA3's own default
differs by at most 1\%) and FlashInfer~\cite{flashinfer} at full GPU clock.
The fused kernel wins on every shape except one: at TP~1 it runs at
0.80--0.92$\times$ FlashInfer's time and 0.56--0.85$\times$ FA3's; on a TP~2
rank it runs at 0.79--0.97$\times$ FlashInfer on 11 of 12 shapes (only
1$\times$32K rises to 1.23$\times$, where FlashInfer stays flat from 8K).
Running the same kernel over raw FragPack records takes 0.90--0.96$\times$
FlashInfer on multi-request TP~1 batches, showing that speedups come from both
kernel design and reading 30\% fewer KV bytes.

Under sustained AgentX serving, both H800 GPUs sit at their 700\,W power cap
in over 99\% of busy samples. Under continuous execution (reaching the 700\,W
cap on all larger batches), BF16 keeps its lead: the fused kernel takes
0.84--0.95$\times$ FlashInfer's time at TP~1 (FA3 takes 1.03--1.19$\times$)
and 0.82--0.98$\times$ at TP~2, except 1$\times$32K (1.35$\times$). Aligning each record's shared-memory
slot with its source address modulo~128 cuts the fused kernel's time at the
cap by 3--6\%, with bit-identical outputs. On FP8 KV, however, the fused kernel
loses its lead at the cap (0.98--1.15$\times$ FlashInfer and
1.21--1.30$\times$ FA3 above 32K tokens). Unfused prefill decompression adds
at most 3\% over FA3 alone.

\subsection{Q3: Compressed cache tiers}
\label{sec:eval-agentx}
A compressed cache tier behaves like a raw tier with $1.4\times$ its
capacity—the inverse of the 0.71 stored ratio. Between raw's LRU cliff and
compressed's cliff, tier mode delivers 2.7--3.6$\times$ raw's goodput; where
raw already fits the working set, tier mode matches it.

\begin{table}[t]
\centering
\caption{AgentX on Qwen3-30B-A3B with 400\,GB of DRAM and 64 Sessions, by SSD Size}
\label{tab:agentx}
\setlength{\tabcolsep}{3.6pt}
\begin{tabular}{@{}llrrrrrr@{}}
\toprule
& & Good- & Hit & \multicolumn{2}{c}{TTFT (s)} & ITL & k tok \\
\cmidrule(lr){5-6}
SSD & Configuration & put & rate & mean & p99 & (ms) & /s \\
\midrule
2\,TB & Raw & 15.1 & 0.94 & 1.69 & 18.6 & 26.0 & 27.5 \\
    & Tier mode & 15.1 & 0.94 & 1.73 & 17.6 & 26.0 & 26.9 \\
640\,GB & Raw & 4.6 & 0.55 & 6.76 & 78.7 & 43.5 & 12.2 \\
    & Tier mode & \textbf{13.5} & 0.91 & 2.31 & 41.6 & 28.4 & 24.6 \\
480\,GB & Raw & 3.5 & 0.44 & 7.66 & 73.0 & 53.9 & 11.1 \\
    & Tier mode & \textbf{9.3} & 0.81 & 3.97 & 42.3 & 34.6 & 19.4 \\
\midrule
\multicolumn{8}{@{}l}{\emph{Other configurations with 640\,GB of SSD}} \\
    & Tier, fixed slots & 8.4 & 0.83 & 3.92 & 34.2 & 35.7 & 19.5 \\
    & FP8 KV, raw & 20.6 & 0.94 & 1.06 & 6.6 & 21.6 & 33.0 \\
    & FP8 KV, tier mode & 20.6 & 0.94 & 1.02 & 4.8 & 21.7 & 33.1 \\
\midrule
\multicolumn{8}{@{}l}{\emph{DRAM only, 40 sessions}} \\
    & Raw & 2.9 & 0.45 & 7.33 & 48.4 & 62.0 & 11.3 \\
    & Tier mode & \textbf{12.5} & 0.91 & 1.56 & 15.7 & 30.7 & 25.1 \\
\midrule
\multicolumn{8}{@{}l}{\emph{GLM-4-9B, 400\,GB of SSD}} \\
    & Raw & 3.5 & 0.44 & 6.82 & 71.6 & 56.6 & 11.1 \\
    & Tier mode & \textbf{13.4} & 0.89 & 1.74 & 26.0 & 31.0 & 24.1 \\
\midrule
\multicolumn{8}{@{}l}{\emph{GLM-4-9B, DRAM only, 48 sessions}} \\
    & Raw & 1.9 & 0.43 & 7.38 & 48.3 & 62.8 & 10.1 \\
    & Tier mode & \textbf{14.4} & 0.92 & 1.53 & 11.3 & 29.0 & 27.4 \\
\bottomrule
\end{tabular}\\[2pt]
\parbox{\linewidth}{\footnotesize $S$ = 8, over the
profiling window. Goodput: requests per minute within the latency targets; hit rate: share of prompt
tokens served from a cache; k tok/s: throughput. Bold: the higher goodput of each raw/compressed pair, marked only when the gap exceeds the variation between repeated runs.}
\end{table}

\begin{figure}[t]
\centering
\includegraphics{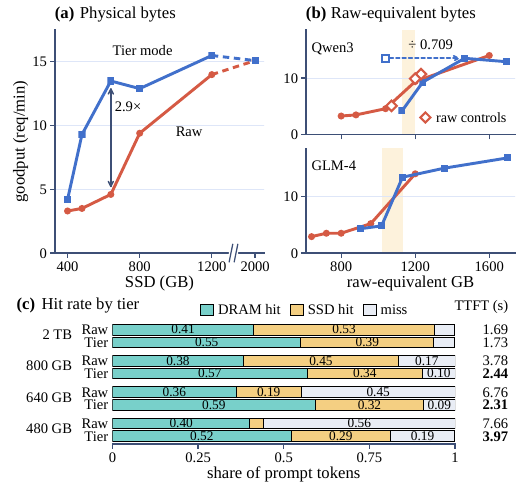}
\caption{AgentX with 64 sessions and 400\,GB of DRAM. (a)~Goodput by physical
SSD size on Qwen3-30B-A3B. (b)~Goodput replotted by raw-equivalent DRAM plus
SSD bytes (tier mode divides physical bytes by the stored ratio, shown by the
arrow for 640\,GB). Top: Qwen3-30B-A3B, with hollow diamonds marking three
enlarged-tier raw controls. Bottom: GLM-4-9B (240--800\,GB SSD). Shaded
bands mark the cliff interval. (c)~Hit rate split by serving tier (DRAM vs.\
SSD vs.\ miss) alongside mean engine TTFT.}
\label{fig:equiv}
\end{figure}

\textbf{SSD capacity sweep and cliff dynamics (Fig.~\ref{fig:equiv}a, Table~\ref{tab:agentx}).}
Sweeping SSD capacity from 2\,TB down to 400\,GB shifts the LRU cliff
leftward by $\approx 1.4\times$: raw LMCache falls off its cliff below
1200\,GB while tier mode sustains near-peak goodput down to 640\,GB. Between
the two cliffs, tier mode delivers 2.8--2.9$\times$ raw's goodput at 640\,GB
and 2.7--3.6$\times$ across repeats at 480\,GB, cutting mean engine TTFT by
1.9--3.1$\times$ (Table~\ref{tab:agentx} reports representative points; Fig.~\ref{fig:equiv} plots the full sweep). Inside raw's cliff at 640\,GB, tier mode holds its hit rate steady near 0.90 across the 30-minute window whereas raw decays continuously to 0.17 as prefixes evict shortly before reuse; recomputing 45\% of prompt tokens pushes raw's p90 TTFT from 2.8\,s to 19\,s (tier mode: 3.4\,s) and ITL from 26 to 44\,ms, leaving 63\% of raw requests within the 2\,s target versus 79\% for tier mode. Outside the cliff region, gains taper as expected
(1.2--1.4$\times$ at 800\,GB where raw recovers, and 1.3$\times$ at 400\,GB
where neither holds the working set). The gain is insensitive to the
service-level objective (SLO) definition: at 640\,GB, tier mode serves 2.5--2.9$\times$ raw's goodput across
TTFT targets from 1\,s to 10\,s, and 2.3--2.4$\times$ with no ITL target. On
GLM-4-9B (whose KV is 0.83 of Qwen3's bytes/token), the cliff occurs at
smaller SSD sizes, yielding 3.8$\times$ raw's goodput at 400\,GB
(Table~\ref{tab:agentx}). Storing BF16 KV at 0.71 of its raw size also cuts
the bytes each chunk moves over PCIe and writes to the SSD by 29\%, so every
byte written to the drive holds $1.4\times$ as much KV.

\textbf{Capacity equivalence (Fig.~\ref{fig:equiv}b).} Replotting every run by
raw-equivalent DRAM plus SSD bytes (dividing tier mode's physical bytes by
0.709 on Qwen3 and 0.707 on GLM-4) collapses both modes onto a single curve on
both models. Three enlarged-tier raw controls (hollow diamonds) verify this
directly: tier mode at 880\,GB physical (1241 raw-equivalent GB, serving
9.3--12.5\,req/min across repeats) matches a proportionally enlarged 1231\,GB
raw control (10.7\,req/min), and enlarging only SSD (1072\,GB) or only DRAM
(1200\,GB) falls on the exact same curve. Total DRAM plus SSD governs even
under write-through because the two tiers evict independently and DRAM hits do
not refresh SSD timestamps, allowing DRAM to retain objects already evicted
from SSD.

\textbf{Faster-tier promotion (Fig.~\ref{fig:equiv}c).} Splitting hits by
serving tier reveals two effects. First, raw collapses at the cliff because
SSD hits vanish as objects evict before reuse, driving mean TTFT up sharply.
Second, expanding $L_1$ DRAM by $\approx 1.4\times$ promotes hits upward from
SSD into DRAM at every SSD size (raising DRAM's share from 0.36--0.41 to
0.52--0.59), lowering median engine TTFT even at 2\,TB (0.99\,s to 0.89\,s).
In a controlled 48-prompt repeat trace (8K--128K tokens, 261\,GB raw vs.\
188\,GB compressed), this SSD-to-DRAM promotion cuts second-pass TTFT by
1.6--1.7$\times$ at 220\,GB DRAM (matching 400\,GB raw; 1.3--1.4$\times$ at
160\,GB) while keeping cold first-pass latency within 2\% from 32K upward.

\begin{figure}[t]
\centering
\includegraphics{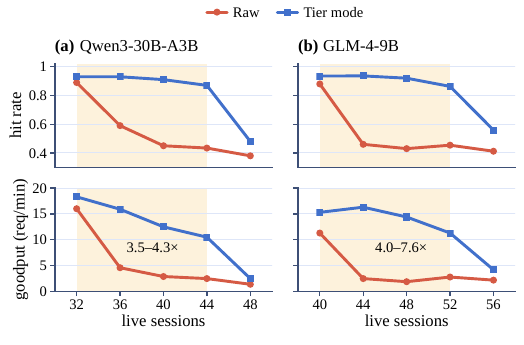}
\caption{DRAM-only deployment (400\,GB DRAM, no SSD): hit rate (top) and
goodput (bottom) versus live sessions $P$ on (a)~Qwen3-30B-A3B and
(b)~GLM-4-9B. Shaded bands mark session counts between raw's cliff and tier
mode's cliff.}
\label{fig:dram}
\end{figure}

\textbf{Single-tier DRAM-only sweep (Fig.~\ref{fig:dram}).} Removing SSD
reproduces the same equivalence: varying live sessions $P$ on 400\,GB DRAM
shifts the sustainable session count by 1.2--1.5$\times$ (centered on
1.4$\times$). Across the shaded cliff bands, tier mode delivers
3.5--4.3$\times$ raw's goodput on Qwen3-30B-A3B and up to 7.6$\times$ on
GLM-4-9B (Table~\ref{tab:agentx}), with only gradual tapering before its own
cliff as recomputed prompt share inches from 0.07 to 0.13.

\textbf{Exact-length ablation and lossy comparison (Table~\ref{tab:agentx}).}
Turning saved bytes into capacity requires exact-length objects. On AgentX KV, TileStream records average 0.710 of raw
size (max 0.843), letting $L_1$ hold $1.41\times$ as many objects; replacing
exact-length allocation with 0.90-size fixed slots shrinks $L_1$ gain to
$1.11\times$ and drops 640\,GB goodput from 13.5 to 8.4\,req/min. Where
accuracy loss is acceptable, lossy FP8 KV halves KV size and reaches
20.6\,req/min at 640\,GB (at the cost of altered outputs). Because \sys{} is
lossless, it composes directly on top of FP8 KV: under DRAM-only serving at
$P=64$, raw FP8 KV overflows 400\,GB DRAM (hit rate 0.61--0.75, goodput
6.6--10.6\,req/min), and enabling tier mode lifts hit rate to 0.92 and goodput
to 19.4\,req/min.

\subsection{Q4: A compressed GPU pool}
\label{sec:eval-hbm}
Once external tiers hold the full working set, GPU KV capacity becomes the
sole bottleneck (Fig.~\ref{fig:motiv}b). Pool mode fits 1.31--1.34$\times$ as
many KV tokens in the same GPU budget (Table~\ref{tab:hbm}), letting a node
admit 12 concurrent long requests with roughly the queue and goodput of a raw
pool at 8, while serving 1.16$\times$ as many prompt tokens per second.

\begin{table}[t]
\centering
\caption{AgentX by GPU KV Budget and Admission Limit $S$}
\label{tab:hbm}
\setlength{\tabcolsep}{5.0pt}
\begin{tabular}{@{}rlrrrrrr@{}}
\toprule
& & Good- & \multicolumn{2}{c}{TTFT (s)} & ITL & & k tok \\
\cmidrule(lr){4-5}
$S$ & Configuration & put & mean & p99 & (ms) & Wait & /s \\
\midrule
\multicolumn{8}{@{}l}{\emph{Qwen3-30B-A3B, 54\,GB/GPU (1.18M vs.\ 1.59M tokens)}} \\
12 & Raw & 10.6 & 5.16 & 43.5 & 34.3 & 1.5 & 28.7 \\
   & Pool mode & \textbf{15.7} & 1.76 & 13.3 & 34.6 & 0.4 & 31.9 \\
16 & Raw & 3.7 & 13.3 & 87.3 & 38.0 & 4.4 & 28.5 \\
   & Pool mode & \textbf{9.1} & 4.43 & 36.8 & 39.8 & 1.4 & 33.5 \\
\midrule
\multicolumn{8}{@{}l}{\emph{Qwen3-30B-A3B, 28\,GB/GPU (0.61M vs.\ 0.80M; larger 0.82M)}} \\
8 & Tier mode & 6.5 & 9.14 & 68.2 & 21.2 & 2.5 & 24.9 \\
  & Pool mode & \textbf{11.9} & 4.63 & 41.2 & 23.1 & 1.3 & 28.3 \\
  & Tier, larger pool & 11.5 & 5.09 & 51.5 & 24.3 & 1.3 & 25.7 \\
\midrule
\multicolumn{8}{@{}l}{\emph{GLM-4-9B, 54\,GB/GPU (1.42M vs.\ 1.90M tokens)}} \\
12 & Raw & 14.1 & 2.39 & 25.4 & 34.7 & 0.5 & 30.9 \\
   & Pool mode & \textbf{19.0} & 1.47 & 9.1 & 30.5 & 0.3 & 35.7 \\
16 & Raw & 6.4 & 7.46 & 69.0 & 41.6 & 2.5 & 31.4 \\
   & Pool mode & \textbf{15.9} & 2.08 & 21.5 & 38.0 & 0.6 & 37.8 \\
\bottomrule
\end{tabular}\\[2pt]
\parbox{\linewidth}{\footnotesize 400\,GB DRAM, 2\,TB SSD, 64 sessions; hit rate is 0.93--0.94 in every row ($S=8$ at 54\,GB is in Fig.~\ref{fig:pool} and Table~\ref{tab:agentx}). Wait: mean waiting requests in vLLM. Raw and tier mode use a raw pool. Bold as in Table~\ref{tab:agentx}.}
\end{table}

\begin{figure}[t]
\centering
\includegraphics{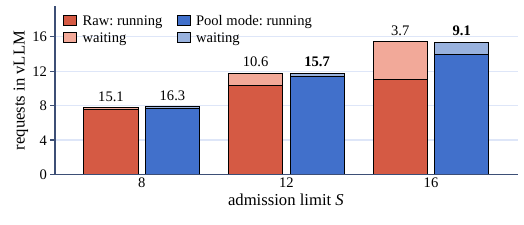}
\caption{Requests running and waiting in vLLM by admission limit $S$, placing
pool mode beside the raw pool of Fig.~\ref{fig:motiv}b (Qwen3-30B-A3B,
54\,GB KV per GPU); bar labels give goodput (req/min).}
\label{fig:pool}
\end{figure}

\textbf{Raising concurrency with ample tiers (Fig.~\ref{fig:pool},
Table~\ref{tab:hbm}).} All runs in Table~\ref{tab:hbm} use 400\,GB DRAM and
2\,TB SSD (0.93--0.94 hit rate throughout), isolating vLLM's internal waiting
queue. At $S=8$, both pools run without queueing and serve similar goodput.
Once admission rises to $S=12$, the raw pool runs out of pages—tripling mean
TTFT and dropping goodput sharply—whereas pool mode keeps waiting flat,
delivers 1.5$\times$ raw's goodput (matching raw's best result at $S=8$), and
lifts prompt throughput by 1.16$\times$ over raw at $S=8$ (1.11$\times$ over raw at $S=12$). At $S=16$, both pools queue, yet pool
mode runs 14.0 concurrent requests against raw's 11.1 and delivers
2.5$\times$ its goodput. GLM-4-9B exhibits the same shift one step higher,
achieving 1.22$\times$ raw's $S=12$ throughput at $S=16$ (Table~\ref{tab:hbm}).

\textbf{Tighter GPU budget, page efficiency, and short tiers.} Shared-budget
pages yield $1.38\times$ raw blocks—versus $1.33\times$
if reserving the worst-case 2\,KB per record and $1.42\times$ for ideal
zero-slack packing. Deducting fixed backend buffers (staging buffers, overflow
area, and record map) from the KV budget trims net token capacity only
modestly, leaving $1.34\times$ usable tokens at 54\,GB/GPU ($\approx 2\text{--}3\%$
buffer share) and $1.31\times$ at 28\,GB/GPU ($\approx 4\text{--}5\%$ share).
At 28\,GB of KV per GPU, even $S=8$ queues inside vLLM. Pool mode halves
waiting and improves goodput by 1.8$\times$ over tier mode's raw pool, matching
a raw pool enlarged to the same token count (\texttt{Tier, larger pool} in
Table~\ref{tab:hbm}) and confirming GPU-pool capacity equivalence. Raw LMCache
is omitted at 28\,GB because its runs stalled when a few loaded-but-unstarted
long requests held 72--82\% of the pool with no running request to preempt—a
vLLM small-pool scheduling artifact. Finally, because pool mode exports
FragPack records unchanged to $L_1$ and $L_2$, it also
relieves short tiers: at 640\,GB SSD ($S=8$), it matches tier mode's goodput
(14.5 vs.\ 13.5\,req/min) with 15\% lower mean TTFT.

\subsection{Q5: A second workload}
\label{sec:eval-ide}
\begin{table}[t]
\centering
\caption{AI IDE workload on 128\,GB of DRAM and 512\,GB of SSD (Qwen3-30B-A3B, TP\,2). Hit rate: share of prompt tokens served by the GPU pool + by the tiers.}
\label{tab:ide}
\footnotesize
\setlength{\tabcolsep}{3.2pt}
\begin{tabular}{@{}lccccc@{}}
\toprule
 & Hit rate & \multicolumn{2}{c}{TTFT (s)} & Queue & Good- \\
Configuration & pool + tiers & mean & p90 & (s) & put \\
\midrule
Raw & 0.58 + 0.22 & 21.9 & 83.4 & 20.2 & 9.4 \\
Tier mode & 0.64 + 0.22 & 3.9 & 11.2 & 2.6 & 11.6 \\
Pool mode & 0.81 + 0.06 & 1.8 & 4.1 & 0.65 & 15.0 \\
\midrule
Raw, 400\,GB + 2\,TB & 0.87 (total) & 2.7 & 7.4 & 1.5 & 12.7 \\
\bottomrule
\end{tabular}
\end{table}

On a second workload where both the external tiers and the GPU pool are
constrained, tier mode removes most of raw's excess TTFT, and pool mode
removes most of the remainder.

This trace replays an internal AI IDE service open-loop over three hours
(5,098 requests at 0.47\,req/s; mean prompt 68K, p99 192K; mean output 505
tokens) with no FIFO admission proxy, so TTFT directly includes vLLM queueing
(Table~\ref{tab:ide}). On a constrained node (128\,GB DRAM + 512\,GB SSD), raw
LMCache loses 7 percentage points of hit rate relative to a large raw reference
(0.80 vs.\ 0.87 at 400\,GB + 2\,TB); recomputing those long prefixes saturates
the GPUs, inflating mean queueing past 20\,s and pushing ITL to 54\,ms. Tier
mode nearly restores total hit rate (0.86 vs.\ 0.87)—cutting mean TTFT by
5.6$\times$ and lifting goodput by 1.24$\times$—but requests still queue as the
GPU pool becomes the bottleneck. Pool mode relieves both levels at once: expanding GPU pool tokens
shifts most hits directly into the GPU pool, slashing queueing wait and
raising goodput by 1.59$\times$ over raw—outperforming even the large raw
reference despite having one-third the DRAM and one-quarter the SSD
(Table~\ref{tab:ide}).

\subsection{Q6: Costs and exactness}
\label{sec:eval-cost}
Where raw tiers already hold the working set, compression costs at most 2.2\%
of throughput. Tier mode produces outputs bit-identical to raw LMCache, while
pool mode stores every KV bit losslessly and matches FA3's numerical
accuracy.

\textbf{Overheads.} Four system costs bound the price of compression:
(1)~\emph{GPU memory:} tier mode uses about 2.5\,GB more GPU memory per card
than raw LMCache (2.1\,GB for 32-chunk store and rotating load buffers).
(2)~\emph{Throughput with ample tiers:} at equal hit rate and goodput, tier
mode costs 1.3\% throughput on DRAM-only ($P=24$) and 2.2\% on 2\,TB SSD.
(3)~\emph{Zero-reuse store cost:} on 320 unique 32K-token prompts with zero
hits, end-to-end time grows by 0.3\% in tier mode and 0.9\% in pool mode.
(4)~\emph{GPU interference and compute share:} running tier mode's store and
load paths at AgentX's chunk rate beside an offline vLLM job slows vLLM by
1.5\% (vs.\ 0.5\% for raw copies) due to multi-process time slicing; in pool
mode, FragPack compression takes 1.5\% of GPU time against 48\% for attention.

\textbf{Tier mode: bit-exact outputs.} GPU unit tests verify every
decompressed block byte for byte across raw fallback, partial tiles, and
corrupted headers. End to end, greedy decoding outputs with compressed tiers
match raw-tier outputs by SHA-1 hash on both Qwen3-30B-A3B and GLM-4-9B—key
for reproducible evaluation, reinforcement-learning (RL) rollouts~\cite{fp8rl,kv4rl}, and
safety-sensitive deployments where KV quantization degrades
alignment~\cite{aligncollapse}.

\textbf{Pool mode: lossless storage with a fused kernel.} The fused decode
kernel produces bitwise identical outputs over compressed and raw FragPack
records, confirming zero coding error. Because the fused kernel accumulates
attention outputs in a different floating-point summation order from FA3,
final logits differ only at rounding precision: against an FP64 reference,
mean absolute decode error matches FA3 ($3.1 \times 10^{-5}$ vs.\
$3.1 \times 10^{-5}$ on BF16; $3.0 \times 10^{-5}$ vs.\ $5.1 \times 10^{-4}$
on FP8, where FA3 also quantizes queries). End-to-end accuracy agrees on
GSM8K~\cite{gsm8k} and 32K--128K RULER~\cite{ruler} multi-key needles and
five-hop variable tracking (within one standard error, paired McNemar
$p \ge 0.42$), whereas vLLM's lossy FP8 KV~\cite{vllmfp8} misses needles that
FA3 retrieves ($p = 0.03$, uncorrected).

%% file: sections/09-mode-choice.tex
\section{Discussion}
\label{sec:disc}
Both modes of \sys{} share the same lossless code; deciding which mode to
enable depends on which level of the KV hierarchy exhausts capacity first, as
observed from live serving metrics.

\textbf{Choosing a mode.} Our evaluation points to four operating regimes.
(1)~\emph{External tiers are the bottleneck: use tier mode.} Observable signs
are frequent SSD hits or a cache hit rate noticeably below that of unbounded
tiers. Tier mode makes existing DRAM and SSD behave like $1.4\times$ larger
raw tiers—delivering its largest gains near the LRU cliff—while preserving
FA3's exact output bits so model outputs need no re-qualification.
(2)~\emph{The GPU pool is the bottleneck: use pool mode.} Observable signs
are a high cache hit rate alongside a persistent waiting queue inside the
inference engine. Here external tiers already fit the working set, so only
compressing the GPU pool increases active concurrency; the fused decode kernel
requires one-time validation, matching FA3 up to floating-point rounding.
(3)~\emph{Both levels are short: use pool mode.} Because pool mode exports
its FragPack records unchanged to $L_1$ and $L_2$, it expands both the GPU
pool and the external tiers simultaneously, delivering the highest gains when
both levels are constrained, as on our AI IDE workload.
(4)~\emph{Neither level is short: disable compression.} When raw tiers and
the raw pool already hold the working set without queueing, compression adds
no capacity benefit and costs up to 2.2\% of throughput. Across all cases,
compression pays off most where physical capacity cannot simply be added—GPU
memory, DRAM-only nodes, and nodes with fixed local NVMe—while saving storage
hardware wherever SSD capacity is elastic.

\textbf{Capacity sizing rule.} Operators can size external tiers in two
steps. First, estimate the active reused working set $W$. Under LMCache's
write-through LRU, raw deployments clear the cliff once total DRAM plus SSD
reaches roughly $1.8W$, or once DRAM alone reaches $1.2\text{--}1.3W$. Both
multiples exceed $W$ because $W$ measures only the mean KV of live sessions,
whereas LRU also retains expired chunks until they age out, and write-through
duplicates hot chunks across both DRAM and SSD. Second, multiply this raw
requirement by \sys{}'s stored ratio: compressed tiers need 0.71 of the raw
DRAM and SSD capacity for BF16 KV, and 0.85--0.87 for FP8 KV.

\textbf{Composing with replacement policies.} Compression reduces the bytes
per cached prefix without altering which prefixes a replacement policy
chooses to retain, so the two compose orthogonally. To verify this, we replay
the measured AgentX request traces through a cache simulator that models
compression by scaling tier capacity by the inverse stored ratio
($1 / 0.709$); this reproduces both measured LRU cliffs within 80\,GB.
Replacing LRU in the simulator confirms composition across policies. On DRAM
alone, LRU over compressed records caches 0.81--0.92 of prompt tokens between
the cliffs, surpassing even Belady's optimal offline policy~\cite{belady} over
raw bytes (0.78--0.88). With DRAM plus SSD, replacing write-through LRU with a
single exclusive LRU~\cite{wongwilkes} (keeping each object in only one tier)
shifts the raw cliff to fewer bytes—caching at most 0.29 of prompt tokens up
to 720\,GB and jumping to 0.88 at 840\,GB—and compression shifts the
exclusive-LRU cliff by the exact same 0.71 factor (reaching 0.87 hit rate at
600\,GB). Thus, the replacement policy sets the baseline cliff location, and
lossless compression consistently scales it down to 0.71 of the bytes.

\textbf{Scope and limitations.} Our evaluation covers single-node serving at
TP~1 and TP~2 over PCIe. Each TP rank compresses, stores, and decompresses
its own KV heads with their own per-head codebooks and no cross-GPU
communication, so the stored ratio does not depend on the TP degree. Pipeline and expert
parallelism, multi-engine shared cache servers, multi-head latent attention
(MLA), and FP4 KV remain untested;
for MLA, TileStream can compress latent vectors directly, whereas pool mode's
GQA kernel would need adaptation. In hybrid architectures mixing attention
with linear or state-space layers, \sys{} compresses only attention KV
layers. Pool mode's current backend supports BF16 and FP8 KV, 128-dimension
heads, GQA group sizes up to~8, and 128-token blocks. Finally, FP8 pool mode
yields limited end-to-end gain for two reasons: shared pages must be sized
for the hardest-to-compress layer (0.906 ratio versus 0.838 mean), and the
FP8 fused kernel runs slower than FA3 at the power cap.

%% file: sections/11-related-work.tex
\section{Related Work}
\label{sec:related}
\textbf{Tiered KV caches.} LMCache~\cite{lmcache}, Mooncake~\cite{mooncake},
CachedAttention~\cite{cachedattention}, SGLang HiCache~\cite{hicache}, and
Strata~\cite{strata} offload uncompressed KV to DRAM and SSD to avoid
repeated prefill. Tutti~\cite{tutti} accelerates SSD loading via
GPU-initiated IO, whereas \sys{} increases how many prefixes fit in each
tier. Placement~\cite{kvplace} and replacement
policies~\cite{arc,wongwilkes,belady} decide which chunks to retain, whereas
lossless compression shrinks each chunk and composes orthogonally with them.

\textbf{Lossy KV compression.} Quantization schemes such
as CacheGen~\cite{cachegen}, KVTC~\cite{kvtc}, KIVI~\cite{kivi},
KVQuant~\cite{kvquant}, FP8 KV~\cite{vllmfp8}, and PackKV~\cite{packkv}, as
well as token-eviction methods such as H$_2$O~\cite{h2o} and
SnapKV~\cite{snapkv}, save more space than lossless coding but alter KV
values and model outputs; \sys{} preserves every bit and can also layer
losslessly on top of quantized formats such as FP8 KV.

\textbf{Lossless numerical compression.} Beyond general-purpose GPU compressors
(nvCOMP~\cite{nvcomp}), offline weight compressors
(DFloat11~\cite{dfloat11}, ZipNN~\cite{zipnn}, ZipServ~\cite{zipserv}, and
Heilper and Singer~\cite{lowprecnn}), and transfer-oriented
SplitZip~\cite{splitzip} discussed in \S\ref{sec:bg}, several concurrent
systems target lossless KV coding. ExANS~\cite{exans} encodes paged BF16
exponents with ANS (Table~\ref{tab:codec}), achieving a slightly lower ratio
than \sys{} but decompressing 6.3$\times$ slower and lacking fixed lane offsets
for fused attention. AdaCode~\cite{adacode} achieves a 0.72 average BF16 KV
ratio. From database storage, PFOR~\cite{pfor} packs fixed-width codes with
out-of-band exceptions, ALP-RD~\cite{alp} pairs a 3-bit leading-bit
dictionary with exceptions, and Parquet's byte-stream-split~\cite{parquet}
separates bytes by position as TileStream does; \sys{} combines a two-level
3-bit escape code with a lane-parallel GPU layout akin to
FastLanes~\cite{fastlanes}.

\textbf{Fused attention over compressed KV.}
Quantized attention kernels dequantize on the fly, and ZipServ fuses
decompression into weight GEMMs. Concurrent ZipFusion~\cite{zipfusion}
decompresses BF16 KV inside FlashInfer decode attention at a 0.79 ratio
(Table~\ref{tab:codec}) without external tiers. Concurrent GAAC~\cite{gaac}
targets personal devices with calibrated per-channel exponent codes
(0.712 stored ratio, i.e., 1.405$\times$ capacity, on BF16), a fused kernel,
and fixed slots; \sys{} targets multi-request serving engines, pools
variable-length budgets across page records, integrates with tiered storage,
and tightens FP8 KV (0.85--0.87 vs.\ 0.88--0.92).

\textbf{Compressed buffer management.} OS compression caches
such as Douglis~\cite{douglis93} and zswap~\cite{zswap} interpose a
compressed tier ahead of slower storage, flash caches such as
CacheLib~\cite{cachelib} and Kangaroo~\cite{kangaroo} manage variable-length
SSD objects, and database engines such as InnoDB~\cite{innodb} and
Umbra~\cite{umbra} pack or resize compressed pages; \sys{} brings
exact-length tier allocation and shared-budget compressed GPU pages to
unmodified LLM serving stacks.

%% file: sections/12-conclusion.tex
\section{Conclusion}
We presented \sys{}, a lossless KV cache compression system spanning both
external cache tiers (DRAM and SSD) and the GPU KV pool without modifying the
inference engine. Built on a calibrated two-level 3-bit high-byte code,
\sys{} stores BF16 KV at 0.71 of its raw size across nine models and turns
those savings into capacity via exact-length tier objects (TileStream) and
shared-budget compressed GPU pages (FragPack), paired with branchless
direct-to-page decompression, fused decode attention, and stall-free
asynchronous transfers. On AgentX, compressed tiers act as raw tiers with
$1.4\times$ the capacity—delivering 2.7--3.6$\times$ raw's goodput across
LMCache's LRU cliff—while a compressed GPU pool admits 12 long requests with
roughly the queue and goodput of a raw pool at 8. On an AI IDE workload where
both tiers and the GPU pool are constrained, tier mode and pool mode lift
goodput by 1.24$\times$ and 1.59$\times$, respectively. Where raw capacity already suffices, compression
costs at most 2.2\% of throughput; tier mode preserves bit-identical outputs,
and pool mode matches FA3 up to floating-point rounding.

%% file: main.bbl
\begin{thebibliography}{10}
\providecommand{\url}[1]{#1}
\csname url@samestyle\endcsname
\providecommand{\newblock}{\relax}
\providecommand{\bibinfo}[2]{#2}
\providecommand{\BIBentrySTDinterwordspacing}{\spaceskip=0pt\relax}
\providecommand{\BIBentryALTinterwordstretchfactor}{4}
\providecommand{\BIBentryALTinterwordspacing}{\spaceskip=\fontdimen2\font plus
\BIBentryALTinterwordstretchfactor\fontdimen3\font minus
  \fontdimen4\font\relax}
\providecommand{\BIBforeignlanguage}[2]{{%
\expandafter\ifx\csname l@#1\endcsname\relax
\typeout{** WARNING: IEEEtran.bst: No hyphenation pattern has been}%
\typeout{** loaded for the language `#1'. Using the pattern for}%
\typeout{** the default language instead.}%
\else
\language=\csname l@#1\endcsname
\fi
#2}}
\providecommand{\BIBdecl}{\relax}
\BIBdecl

\bibitem{kvwild}
J.~Wang \emph{et~al.}, ``{KVCache} cache in the wild: Characterizing and
  optimizing {KVCache} cache at a large cloud provider,'' in \emph{Proc. USENIX
  Annu. Tech. Conf. (ATC)}, 2025.

\bibitem{copilot}
B.~Liu \emph{et~al.}, ``Agentic coding in the wild: Characterizing {GitHub
  Copilot} traces at production scale,'' arXiv:2608.00101, 2026.

\bibitem{vllm}
W.~Kwon \emph{et~al.}, ``Efficient memory management for large language model
  serving with {PagedAttention},'' in \emph{Proc. 29th Symp. Operating Systems
  Principles (SOSP)}, 2023, pp. 611--626.

\bibitem{sglang}
L.~Zheng \emph{et~al.}, ``{SGLang}: Efficient execution of structured language
  model programs,'' in \emph{Proc. Advances in Neural Information Processing
  Systems (NeurIPS)}, 2024.

\bibitem{lmcache}
Y.~Liu \emph{et~al.}, ``{LMCache}: An efficient {KV} cache layer for
  enterprise-scale {LLM} inference,'' arXiv:2510.09665, 2025.

\bibitem{mooncake}
R.~Qin \emph{et~al.}, ``Mooncake: Trading more storage for less computation---a
  {KVCache}-centric architecture for serving {LLM} chatbot,'' in \emph{Proc.
  23rd USENIX Conf. File and Storage Technologies (FAST)}, 2025.

\bibitem{hicache}
\BIBentryALTinterwordspacing
{SGLang Team}, ``{SGLang HiCache}: Fast hierarchical {KV} caching with your
  favorite storage backends,'' LMSYS blog, 2025. [Online]. Available:
  \url{https://lmsys.org/blog/2025-09-10-sglang-hicache/}
\BIBentrySTDinterwordspacing

\bibitem{cachedattention}
B.~Gao \emph{et~al.}, ``Cost-efficient large language model serving for
  multi-turn conversations with {CachedAttention},'' in \emph{Proc. USENIX
  Annu. Tech. Conf. (ATC)}, 2024.

\bibitem{strata}
Z.~Xie \emph{et~al.}, ``Strata: Hierarchical context caching for long context
  language model serving,'' in \emph{Proc. 20th USENIX Symp. Operating Systems
  Design and Implementation (OSDI)}, 2026.

\bibitem{kivi}
Z.~Liu \emph{et~al.}, ``{KIVI}: A tuning-free asymmetric 2bit quantization for
  {KV} cache,'' in \emph{Proc. 41st Int. Conf. Machine Learning (ICML)}, 2024.

\bibitem{cachegen}
Y.~Liu \emph{et~al.}, ``{CacheGen}: {KV} cache compression and streaming for
  fast large language model serving,'' in \emph{Proc. ACM SIGCOMM}, 2024, pp.
  38--56.

\bibitem{kvtc}
K.~Staniszewski and A.~{\L}a{\'n}cucki, ``{KV} cache transform coding for
  compact storage in {LLM} inference,'' in \emph{Proc. Int. Conf. Learning
  Representations (ICLR)}, 2026.

\bibitem{h2o}
Z.~Zhang \emph{et~al.}, ``{H$_2$O}: Heavy-hitter oracle for efficient
  generative inference of large language models,'' in \emph{Proc. Advances in
  Neural Information Processing Systems (NeurIPS)}, 2023.

\bibitem{snapkv}
Y.~Li \emph{et~al.}, ``{SnapKV}: {LLM} knows what you are looking for before
  generation,'' in \emph{Proc. Advances in Neural Information Processing
  Systems (NeurIPS)}, 2024.

\bibitem{fa3}
J.~Shah \emph{et~al.}, ``{FlashAttention-3}: Fast and accurate attention with
  asynchrony and low-precision,'' in \emph{Proc. Advances in Neural Information
  Processing Systems (NeurIPS)}, 2024.

\bibitem{flashinfer}
Z.~Ye \emph{et~al.}, ``{FlashInfer}: Efficient and customizable attention
  engine for {LLM} inference serving,'' in \emph{Proc. Conf. Machine Learning
  and Systems (MLSys)}, 2025.

\bibitem{zipserv}
R.~Fan \emph{et~al.}, ``{ZipServ}: Fast and memory-efficient {LLM} inference
  with hardware-aware lossless compression,'' in \emph{Proc. 31st ACM Int.
  Conf. Architectural Support for Programming Languages and Operating Systems
  (ASPLOS)}, 2026, pp. 2264--2280.

\bibitem{agentx}
\BIBentryALTinterwordspacing
{SemiAnalysis}, ``{InferenceX AgentX}: Methodology and datasets,'' 2026.
  [Online]. Available: \url{https://inferencex.semianalysis.com/infx/agentx}
\BIBentrySTDinterwordspacing

\bibitem{nvcomp}
\BIBentryALTinterwordspacing
{NVIDIA}, ``{nvCOMP}: {GPU}-accelerated compression and decompression
  library.'' [Online]. Available: \url{https://developer.nvidia.com/nvcomp}
\BIBentrySTDinterwordspacing

\bibitem{dfloat11}
T.~Zhang \emph{et~al.}, ``70\% size, 100\% accuracy: Lossless {LLM} compression
  for efficient {GPU} inference via dynamic-length float ({DFloat11}),'' in
  \emph{Proc. Advances in Neural Information Processing Systems (NeurIPS)},
  2025.

\bibitem{zipnn}
M.~Hershcovitch \emph{et~al.}, ``{ZipNN}: Lossless compression for {AI}
  models,'' in \emph{Proc. IEEE 18th Int. Conf. Cloud Computing (CLOUD)}, 2025,
  pp. 186--198.

\bibitem{splitzip}
Y.~Guo and S.~Joshi, ``{SplitZip}: Ultra fast lossless {KV} compression for
  disaggregated {LLM} serving,'' arXiv:2605.01708, 2026.

\bibitem{giesen}
F.~Giesen, ``Interleaved entropy coders,'' arXiv:1402.3392, 2014.

\bibitem{cctraces}
\BIBentryALTinterwordspacing
{SemiAnalysis}, ``{CC} traces---{Weka}, with subagents, 256k cap,'' 2026.
  [Online]. Available:
  \url{https://huggingface.co/datasets/semianalysisai/cc-traces-weka-062126-256k}
\BIBentrySTDinterwordspacing

\bibitem{alp}
A.~Afroozeh, L.~X. Kuff{\'o}, and P.~Boncz, ``{ALP}: Adaptive lossless
  floating-point compression,'' \emph{Proc. ACM Manag. Data}, vol.~1, no.~4,
  pp. 1--26, 2023.

\bibitem{exans}
\BIBentryALTinterwordspacing
{OpenLake}, ``{ExANS}: Achieving 600\,{GB/s}+ decode throughput at 1.51x
  lossless compression,'' 2026. [Online]. Available:
  \url{https://www.theopenlake.com/blog/exans-lossless-gpu-compression-for-bf16-kv-cache}
\BIBentrySTDinterwordspacing

\bibitem{zipfusion}
Y.~Chen, P.~Wang, and H.~Xu, ``{ZipFusion}: Fusing lossless {KV} decompression
  with attention for efficient {LLM} decoding,'' arXiv:2610.33719, 2026.

\bibitem{shadowserve}
X.~Xiang \emph{et~al.}, ``{ShadowServe}: Interference-free {KV} cache fetching
  for distributed prefix caching,'' arXiv:2509.16857, 2025.

\bibitem{fp8rl}
Z.~Qiu \emph{et~al.}, ``{FP8-RL}: A practical and stable low-precision stack
  for {LLM} reinforcement learning,'' arXiv:2601.18150, 2026.

\bibitem{kv4rl}
S.~Park, S.-Y. Kim, D.~Jang, and E.~Yang, ``{KV4-RL}: Unlocking faster
  asynchronous rollouts for {RLVR} with {INT4} {KV} cache quantization,''
  arXiv:2610.14448, 2026.

\bibitem{aligncollapse}
B.~C. Xu, A.~Kumarappan, and M.~Zhou, ``Alignment collapse under {KV} cache
  quantization: Diagnosis and mitigation,'' arXiv:2606.09864, 2026.

\bibitem{gsm8k}
K.~Cobbe \emph{et~al.}, ``Training verifiers to solve math word problems,''
  arXiv:2110.14168, 2021.

\bibitem{ruler}
C.-P. Hsieh \emph{et~al.}, ``{RULER}: What's the real context size of your
  long-context language models?'' in \emph{Proc. Conf. Language Modeling
  (COLM)}, 2024.

\bibitem{vllmfp8}
\BIBentryALTinterwordspacing
J.~K{\"u}bler \emph{et~al.}, ``The state of {FP8} {KV}-cache and attention
  quantization in {vLLM},'' vLLM Blog, 2026. [Online]. Available:
  \url{https://vllm.ai/blog/2026-04-22-fp8-kvcache}
\BIBentrySTDinterwordspacing

\bibitem{belady}
L.~A. Belady, ``A study of replacement algorithms for a virtual-storage
  computer,'' \emph{IBM Syst. J.}, vol.~5, no.~2, pp. 78--101, 1966.

\bibitem{wongwilkes}
T.~M. Wong and J.~Wilkes, ``My cache or yours? {M}aking storage more
  exclusive,'' in \emph{Proc. USENIX Annu. Tech. Conf. (ATC)}, 2002.

\bibitem{tutti}
S.~Qiu \emph{et~al.}, ``Tutti: Making {SSD}-backed {KV} cache practical for
  long-context {LLM} serving,'' arXiv:2605.03375, 2026.

\bibitem{kvplace}
S.~D. Tumkur \emph{et~al.}, ``Where should the {KV} cache live? {P}lacement
  policies across {GPU}, {CPU}, and {SSD} for long-lived sessions,''
  arXiv:2609.16215, 2026.

\bibitem{arc}
N.~Megiddo and D.~S. Modha, ``{ARC}: A self-tuning, low overhead replacement
  cache,'' in \emph{Proc. 2nd USENIX Conf. File and Storage Technologies
  (FAST)}, 2003.

\bibitem{kvquant}
C.~Hooper \emph{et~al.}, ``{KVQuant}: Towards 10 million context length {LLM}
  inference with {KV} cache quantization,'' in \emph{Proc. Advances in Neural
  Information Processing Systems (NeurIPS)}, 2024.

\bibitem{packkv}
B.~Jiang \emph{et~al.}, ``{PackKV}: Reducing {KV} cache memory footprint
  through {LLM}-aware lossy compression,'' in \emph{Proc. IEEE Int. Parallel
  and Distributed Processing Symp. (IPDPS)}, 2026, pp. 598--611.

\bibitem{lowprecnn}
A.~Heilper and D.~Singer, ``Lossless compression of neural network components:
  Weights, checkpoints, and {K/V} caches in low-precision formats,''
  arXiv:2508.19263, 2025.

\bibitem{adacode}
C.~Li, P.~Liu, L.~Chen, Y.~Zhai, and J.~Z. Yang, ``{AdaCode}: Adaptive lossless
  compression for {BFloat16} {KV} cache,'' in \emph{Proc. Conf. Empirical
  Methods in Natural Language Processing (EMNLP)}, 2026, pp. 6112--6130.

\bibitem{pfor}
M.~Zukowski, S.~H{\'e}man, N.~Nes, and P.~Boncz, ``Super-scalar {RAM-CPU} cache
  compression,'' in \emph{Proc. 22nd Int. Conf. Data Engineering (ICDE)}, 2006,
  p.~59.

\bibitem{parquet}
\BIBentryALTinterwordspacing
{Apache Parquet}, ``Encodings: {BYTE\_STREAM\_SPLIT}.'' [Online]. Available:
  \url{https://parquet.apache.org/docs/file-format/data-pages/encodings/}
\BIBentrySTDinterwordspacing

\bibitem{fastlanes}
A.~Afroozeh and P.~Boncz, ``The {FastLanes} compression layout: Decoding $>$100
  billion integers per second with scalar code,'' \emph{Proc. VLDB Endow.},
  vol.~16, no.~9, pp. 2132--2144, 2023.

\bibitem{gaac}
Y.~Guo, A.~M. George, and S.~Joshi, ``{GAAC}: Lossless {KV} compression for
  bandwidth-efficient {LLM} inference on personal devices,'' arXiv:2610.38210,
  2026.

\bibitem{douglis93}
F.~Douglis, ``The compression cache: Using on-line compression to extend
  physical memory,'' in \emph{Proc. USENIX Winter Conf.}, 1993.

\bibitem{zswap}
\BIBentryALTinterwordspacing
{Linux kernel documentation}, ``zswap.'' [Online]. Available:
  \url{https://docs.kernel.org/admin-guide/mm/zswap.html}
\BIBentrySTDinterwordspacing

\bibitem{cachelib}
B.~Berg \emph{et~al.}, ``The {CacheLib} caching engine: Design and experiences
  at scale,'' in \emph{Proc. 14th USENIX Symp. Operating Systems Design and
  Implementation (OSDI)}, 2020.

\bibitem{kangaroo}
S.~McAllister \emph{et~al.}, ``Kangaroo: Caching billions of tiny objects on
  flash,'' in \emph{Proc. 28th ACM Symp. Operating Systems Principles (SOSP)},
  2021.

\bibitem{innodb}
\BIBentryALTinterwordspacing
{Oracle}, ``{InnoDB} table and page compression,'' MySQL 8.0 Reference Manual.
  [Online]. Available:
  \url{https://dev.mysql.com/doc/refman/8.0/en/innodb-compression.html}
\BIBentrySTDinterwordspacing

\bibitem{umbra}
T.~Neumann and M.~J. Freitag, ``Umbra: A disk-based system with in-memory
  performance,'' in \emph{Proc. 10th Conf. Innovative Data Systems Research
  (CIDR)}, 2020.

\end{thebibliography}
